\documentclass[cha%
 aip,
 amsmath,amssymb,
 reprint,%
]{revtex4-1}

\usepackage{graphicx}% Include figure files
\usepackage{makecell}

\usepackage{dcolumn}% Align table columns on decimal point
\usepackage{bm}% bold math
\usepackage[utf8]{inputenc}
\usepackage[T1]{fontenc}
\usepackage{mathptmx}
\usepackage{etoolbox}
\usepackage{amsthm}

\newtheorem{lemma}{Lemma}

\makeatletter
\newcounter{myalgorithm}
\renewcommand{\themyalgorithm}{\Roman{myalgorithm}}
\makeatother

\makeatletter
\def\@email#1#2{%
 \endgroup
 \patchcmd{\titleblock@produce}
  {\frontmatter@RRAPformat}
  {\frontmatter@RRAPformat{\produce@RRAP{*#1\href{mailto:#2}{#2}}}\frontmatter@RRAPformat}
  {}{}
}%
\makeatother
\begin{document}

\preprint{AIP/123-QED}

\title[Neural Networks for SDEs with Symmetric Jump Noise]{Neural Networks for Tamed Milstein Approximation of SDEs with Additive Symmetric Jump Noise Driven by a Poisson Random Measure}
% Force line breaks with \\
\author{J.H. Ram\'rez-Gonzalez}
 \altaffiliation[KAUST ]{King Abdullah University of Science and Technology}%Lines break automatically or can be forced with \\
\author{Y. Sun}%
 \email{josehermenegildo.ramirezgonzalez@kaust.edu.sa, ying.sun@kaust.edu.sa}
\affiliation{ 
King Abdullah University of Science and Technology%\\This line break forced with \textbackslash\textbackslash
}%

\date{\today}% It is always \today, today,

\begin{abstract}
This work aims to estimate the drift and diffusion functions in stochastic differential equations (SDEs) driven by a particular class of Lévy processes with finite jump intensity, using neural networks. We propose a framework that integrates the Tamed-Milstein scheme with neural networks employed as non-parametric function approximators. Estimation is carried out in a non-parametric fashion for the drift function \( f: \mathbb{Z} \to \mathbb{R} \), the diffusion coefficient \( g: \mathbb{Z} \to \mathbb{R} \). The model of interest is given by
\[
dX(t) = \xi + f(X(t))\, dt + g(X(t))\, dW_t + \gamma \int_{\mathbb{Z}} z\, N(dt,dz),
\]
where \( W_t \) is a standard Brownian motion, and \( N(dt,dz) \) is a Poisson random measure on \( (~\mathbb{R}_{+} ~\times ~\mathbb{Z}~, ~\mathcal{B}~(~\mathbb{R}_{+}~)~\otimes~\mathcal{Z}~,~ \lambda( \Lambda~\otimes~v~)~) \), with \( \lambda, \gamma > 0 \), \( \Lambda \) being the Lebesgue measure on \( \mathbb{R}_{+} \), and \( v \) a finite measure on the measurable space \( (\mathbb{Z}, \mathcal{Z}) \).

% Additionally, within the same framework, we analyze systems of the form
% \[
% dX_t = f(X_t)\,dt + \gamma X_t \int_{\mathbb{Z}} z\, N(dt,dz),
% \]
% where the jump term is multiplicative, allowing the process to capture proportional, state-dependent discontinuities. 

Neural networks are used as non-parametric function approximators, enabling the modeling of complex nonlinear dynamics without assuming restrictive functional forms. The proposed methodology constitutes a flexible alternative for inference in systems with state-dependent noise and discontinuities driven by Lévy processes.
\end{abstract}

\maketitle

\begin{quotation}
We consider stochastic differential equations that include a drift component, a diffusion component driven by Brownian motion, and an additive jump component governed by a Poisson random measure with finite intensity. The jump sizes are assumed to be symmetric about zero and have finite second moments. Since the transition density of the process is generally not available in closed form, we propose a nonparametric estimation framework based on neural networks trained using loss functions that rely on the minimization of the first and second conditional moments of the increments. These loss functions are constructed from a Tamed–Milstein numerical scheme, enabling the simulation of trajectories required for training. The performance of the method is evaluated through numerical experiments under both continuous and jump-driven settings.
\end{quotation}

\section{Introduction}

Stochastic differential equations (SDEs) are essential tools for modeling systems driven by uncertainty and noise. In finance, the well-known Vasicek and Black--Scholes models describe, respectively, the evolution of interest rates and the pricing of European options via diffusion-type dynamics~\cite{vasicek1977equilibrium, black1973pricing}. However, many phenomena exhibit abrupt changes that standard continuous models fail to capture. To address this, jump-diffusion models such as those introduced by~\cite{merton1976option, kou2002jump, duffie2000transform} incorporate sudden discontinuities into the stochastic framework, allowing for more realistic modeling of asset prices, risk, and volatility. Beyond finance, SDEs with or without jumps have been used in diverse fields including ecology, neuroscience, and climate science, where they effectively model population dynamics~\cite{marrec2023solving}, neuronal variability~\cite{faisal2008noise}, and climate fluctuations~\cite{hasselmann1976stochastic}. These broad applications motivate the study and development of general stochastic models and inference techniques capable of handling both continuous and discontinuous sources of randomness.

A central challenge in the study of stochastic differential equations (SDEs) is the estimation of the drift and diffusion coefficients that characterize the underlying dynamics. 

In this work, we focus on the study of a stochastic system with jumps defined by the equation
\begin{equation}\label{equation_red_1}
    dX(t) = \xi + f(X(t))\, dt + g(X(t))\, dW_t + \gamma \int_{\mathbb{Z}} z\, N(dt,dz),
\end{equation}
where \( W_t \) is a standard Brownian motion and \( N(dt,dz) \) is a Poisson random measure defined on \( ( \mathbb{R}_{+} \times \mathbb{Z}, \mathcal{B}(\mathbb{R}_{+}) \otimes \mathcal{Z}, \lambda \Lambda \otimes v) \), with \( \lambda, \gamma > 0 \), \( \Lambda \) the Lebesgue measure on \( \mathbb{R}_{+} \), and \( v \) a finite measure on the measurable space \( (\mathbb{Z}, \mathcal{Z}) \). 

Traditional parametric approaches impose restrictive assumptions on the functional forms of these components, which may lead to model misspecification. In contrast, neural networks have recently emerged as flexible non-parametric tools capable of approximating such functions directly from data. This approach enables the estimation of drift and diffusion terms without imposing rigid structural constraints. Following this perspective, several works have proposed the use of neural networks for inferring the drift and diffusion coefficients in stochastic differential equations (SDEs), including~\cite{zhang2023milstein,kong2020sdenet,fang2022end}. In~\cite{zhang2023milstein}, the strong order of convergence \( \kappa > 0 \) is defined as the exponent for which there exists a constant \( C > 0 \), independent of the discretization step \( h \), such that
\[
\mathbb{E}\left( \left| \hat{X}_H^{h,n} - X(T) \right| \right) \leq C h^\kappa,
\]
where \( X(T) \) denotes the exact solution at final time \( T = nh \), and \( \{\hat{X}_l^{h,n}\}_{l=0,...,n} \) represents the approximation. The authors emphasize the importance of the convergence order of the approximation to the SDE solution. Under the Tamed–Milstein scheme and considering SDEs where the diffusion coefficient is integrated with respect to Brownian motion, they introduce in their Equation~(3) a specific approximation (the MDSDE-Net model) and demonstrate that it converges in the strong sense with order \( \kappa = 1 \). Additionally, they show that the Euler–Maruyama approximation (used in the SDE-Net model~\cite{kong2020sdenet}) attains a strong convergence order of \( \kappa = 0.5 \). The authors compare their model with several state-of-the-art uncertainty quantification methods, including Deep Ensembles~\cite{lakshminarayanan2017simple}, MC-Dropout~\cite{gal2016dropout}, p-SGLD~\cite{li2016preconditioned}, and SDE-Net. They evaluate the performance across four tasks: the impact of convergence order, out-of-distribution (OOD) detection for classification and regression, and misclassification detection. The metrics used include the true negative rate (TNR) at 0.95 true positive rate (TPR), the area under the receiver operating characteristic curve (AUROC), the area under the precision–recall curve (AUPR), and detection accuracy. The authors report superior performance of their algorithm in most of the considered cases and argue that its robustness stems from the convergence order of the proposed method. However, they do not provide examples evaluating how effectively the functions \( f \) and \( g \) are approximated by their algorithm.

Moreover, in the context of stochastic differential equations with jumps, \cite{fang2022end} proposed a non-parametric estimation framework based on neural networks for inferring the drift and diffusion functions in SDEs driven by symmetric \(\alpha\)-stable noise ($0<\alpha<2$). Their approach relies on an Euler–Maruyama-type discretization adapted to the \(\alpha\)-stable case, which is given by:
\begin{equation}\label{alpha_euler_maruyama}
    A_{t + \Delta t} = A_t + f(A_t) \Delta t + g(A_t) \Delta L_t^{(\alpha)},
\end{equation}
where \( \Delta L_t^{(\alpha)} \) is sampled from the symmetric \( \alpha \)-stable distribution with scale \( (\Delta^{(k)})^{1/\alpha} \) and location $0$. The likelihood function for this process is based on the transition probability of moving from \( A_0^{(k)} \) to \( A_1^{(k)} \) after a time interval \( \Delta^{(k)} \):
\[
A_1^{(k)} \sim S_\alpha\left(g(A_0^{(k)})(\Delta^{(k)})^{1/\alpha}, 0, A_0^{(k)} + \Delta^{(k)} f(A_0^{(k)})\right),
\]
where \( S_\alpha \) represents the symmetric \( \alpha \)-stable distribution with scale \( g(A_0^{(k)})(\Delta^{(k)})^{1/\alpha} \) and location \( A_0^{(k)} + \Delta^{(k)} f(A_0^{(k)}) \). The authors implement a two-step learning procedure. In the first step, the drift function \( f \) is estimated by minimizing a mean squared error loss:
\[
\mathcal{L}_{\text{drift}}(\theta) := \frac{1}{N} \sum_{k=1}^{N} \left[A_1^{(k)} - \left(A_0^{(k)} + \Delta^{(k)} f_\theta(A_0^{(k)})\right) \right]^2,
\]
while in the second step, the diffusion function \( g \) is estimated by maximizing a likelihood function derived from the transition density of the \(\alpha\)-stable distribution:
\scriptsize
\[
\mathcal{L}_{\text{diffusion}}(\theta) := -\frac{1}{N} \sum_{k=1}^{N} \log \left[\frac{1}{g_\theta(A_0^{(k)}) \Delta^{1/\alpha}} p_\alpha \left( \frac{A_1^{(k)} - (A_0^{(k)} + \Delta f_\theta(A_0^{(k)}))}{g_\theta(A_0^{(k)}) \Delta^{1/\alpha}} \right) \right],
\]
\normalsize
where \( p_\alpha \) denotes the density of the standard symmetric \(\alpha\)-stable law. In the special case \(\alpha = 1\), the diffusion term admits a closed-form estimator. The authors also present numerical examples to assess the performance of their algorithm in estimating the functions \( f \) and \( g \).

Under the same line of research, we propose a framework that combines neural networks with the Tamed-Milstein scheme for the nonparametric estimation of drift and diffusion functions in SDEs driven by a class of Lévy processes with finite jump intensity. To perform non-parametric inference via neural networks, we first consider a numerical approximation for the solution of equation~(\ref{equation_red_1}). To this end, we adopt the approximation proposed by~\cite{Kumar} in a more general context. Specifically, the authors propose the Tamed-Milstein scheme to approximate solutions of the stochastic differential equation with jumps of the form
\begin{equation}\label{evolution_1}
dX(t) = \xi + f(X(t))\, dt + g(X(t))\, dW_t + \int_{\mathbb{Z}} \gamma(X(t), z)\, N(dt,dz),
\end{equation}
where \( W_t \) is a standard Brownian motion and \( N(dt,dz) \) is a Poisson random measure defined on \( (\mathbb{Z}, \mathcal{Z}, v) \), with intensity measure \( v \) satisfying \( v(\mathbb{Z}) < \infty \). The functions \( f(x) \) and \( g(x) \) are assumed to be \( \mathcal{B}(\mathbb{R}) \)-measurable, while \( \gamma(x,z) \) is assumed to be \( \mathcal{B}(\mathbb{R}) \otimes \mathcal{Z} \)-measurable. Moreover, all three functions \( f(x) \), \( g(x) \), and \( \gamma(x,z) \) are assumed to be twice differentiable with respect to \( x \in \mathbb{R} \).

Considering a partition of the interval \([0, T]\) into \(n\) sub-intervals of length \(h\) such that \(nh = T\). This approximation is given by:\footnotesize
\begin{multline}\label{aproxh}
X^{h,n}_{(l+1)h} = X^{h,n}_{lh} + \Bigg(f^{h}(X^{h,n}_{lh}) - \int_{\mathbb{Z}}\gamma(X^{h,n}_{lh},z)v(dz)\Bigg) h + g(X^{h,n}_{lh})\, \Delta W_{lh} \\
+ \frac{1}{2} g(X^{h,n}_{lh}) g'(X^{h,n}_{lh}) \Big( (\Delta W_{lh})^2 - h \Big) + \sum_{i=1}^{N((lh,(l+1)h],\mathbb{Z})} \gamma(X^{h,n}_{lh},z_i) \\
+ \sum_{i=1}^{N((lh,(l+1)h],\mathbb{Z})} \Big(g(X^{h,n}_{lh}+\gamma(X^{h,n}_{lh},z_i)) - g(X^{h,n}_{lh})\Big)(W_{(l+1)h} - W_{\tau_i}) \\
+ g(X^{h,n}_{lh}) \sum_{i=1}^{N((lh,(l+1)h],\mathbb{Z})} \frac{\partial\gamma(X^{h,n}_{lh},z_i)}{\partial x}(W_{\tau_i}-W_{lh}) \\
+ \sum_{j=1}^{N((lh,(l+1)h],\mathbb{Z})} \sum_{i=1}^{N((lh,(l+1)h],\mathbb{Z})} \Big(\gamma(X^{h,n}_{lh}+\gamma(X^{h,n}_{lh},z_i),z_j) - \gamma(X^{h,n}_{lh},z_j)\Big)
\end{multline}
\normalsize
for any \( l = 0, \dots, n - 1 \), where  \( z_i \) represents the jump sizes and \( \tau_i \) the corresponding jump times within the interval \( (lh, (l+1)h] \). The number of jumps \( N((lh, (l+1)h], \mathbb{Z}) \) follows a Poisson distribution, i.e., \( N((lh, (l+1)h], \mathbb{Z}) \sim \text{Pois}(\lambda h) \). The jump times \( \tau_i \) are such that the increments \( \tau_i - \tau_{i-1} \) follow an exponential distribution with parameter \( h \), and the jump sizes \( z_i \) are drawn from the normalized measure \( \frac{v(dz)}{v(\mathbb{Z})} \), for \( i = 1, \dots, N((lh, (l+1)h], \mathbb{Z}) \) and $f^{h}(x):=\frac{f(x)}{1+hf^2(x)}$. 

As a consequence of Theorems 2.1 and 2.3, and under assumptions \textbf{(A-1)–(A-5)} as stated in \cite{Kumar}, we have that:
(i) Let Assumptions \textbf{(A-1)-(A-5)} hold with \( p \geq 6(\chi + 2) \). Then, the tamed Milstein scheme~(\ref{aproxh}) satisfies: 
\begin{equation}\label{with_jum_adi}
 \mathbb{E}\left[ \left| X_H^{h,n} - X(T) \right|^2 \right] \leq K h^{-1 -\frac{2}{2+\delta} },
\end{equation}
with \( \delta \in \left( \frac{4}{p - 2}, 1 \right) \), where \( K > 0 \) is a constant independent of \( h \) and \(n\).
(ii) Let Assumptions \textbf{(A1)-(A5)} hold with \( \gamma(x,z) \equiv 0 \) and \( p \geq 6(\chi + 2) \). Then, the tamed Milstein scheme~(\ref{aproxh}) satisfies
\begin{equation}\label{without_jum_adi}
\mathbb{E}\left[  \left| X_H^{h,n} - X(T) \right|^q \right] \leq K h^{-q},
\end{equation}
where \( 0 < q < \max\left(2, \frac{p}{3\chi + 6} - 2\right) \), and \( K > 0 \) is a constant independent of \( h \) and \(n\).

Then, in particular by (i) and (ii) the approximation given in~(\ref{aproxh}) has a strong convergence order \( \kappa = 1 \)
when $\gamma(x,z)=0$ and \( \kappa = \frac{1}{2}+\frac{1}{2+\delta} \) with \( \delta \in \left( \frac{4}{p - 2}, 1 \right) \) in the other case.  Given the importance of the convergence order of numerical schemes in approximating the true solutions of SDEs, and in contrast to \cite{fang2022end} who employ a first-order Euler–Maruyama approximation, we adopt a second-order approximation based on the Tamed–Milstein scheme, defined in equation~(\ref{aproxh}) with \( \gamma(x,z) = \gamma z \). To infer the drift coefficient \( f \) and the diffusion coefficient \( g \) in jump-diffusion SDEs given by equation~(\ref{equation_red_1}) using neural networks.

This paper is organized as follows. Section~\ref{sec2} describes a methodology for estimating the drift coefficient \( f \) and the diffusion coefficient \( g \) using neural networks. Section~\ref{Numerical_Studies_AJ} presents numerical examples based on simulated data using the methodology introduced in Section~\ref{sec2}. In Section~\ref{Discussion}, we summarize the main findings and outline potential directions for future research.

Appendix~A presents a procedure for estimating the characteristic function of the approximation~(\ref{aproxh}) in the specific case where \( \gamma(x,z) = \gamma z \). Appendix~B provides numerical examples in which the \textit{approximated density function} associated with~(\ref{aproxh}) is constructed using the results from~\cite{Yang} and the calculations in Appendix~A. This estimation is essential for inferring the parameters \( \lambda \) and \( \gamma \), as well as for implementing the methodology described in Section~\ref{sec2}. Appendix~C presents two algorithms for estimating the parameters \( \lambda \) and \( \gamma \), corresponding to the jump components in equation~(\ref{equation_red_1}).

\section{Methodology}\label{sec2}
In this section, we present a methodology to estimate the drift and diffusion coefficients corresponding to the SDEs defined by equation~(\ref{equation_red_1}). The approach relies on a discrete-time approximation of the solutions, constructed within the Tamed-Milstein framework and given by equation~(\ref{aproxh}) with \(\gamma(x,z)=\gamma z\). Based on this approximation, inference is performed using neural networks. To implement this procedure, we begin by defining the discrete-time scheme as follows. For $t, \Delta t > 0$, we define:
\begin{multline}\label{conmil2}
X_{t+\Delta t} = X_t + \Bigg(f^{\Delta t}(X_t)-\int_{\mathbb{Z}}\gamma z\,v(dz)\Bigg)\, \Delta t + g(X_t)\, \Delta W_t \\
+ \frac{1}{2} g(X_t) g'(X_t) \Big( (\Delta W_t)^2 - \Delta t \Big) + \gamma \sum_{i=1}^{N((t,t+\Delta t],\mathbb{Z})}z_i \\
+ \sum_{i=1}^{N((t,t+\Delta t],\mathbb{Z})} \Big(g(X_t+\gamma z_i)-g(X_t)\Big)(W_{t+\Delta t}-W_{\tau_i}),
\end{multline}
where \( f^{\Delta t} = \frac{f(x)}{1 + \Delta t f^2(x)} \). Under assumptions \textbf{(A-1)–(A-5)} outlined in \cite{Kumar}, the convergence order of the approximation~(\ref{conmil2}) to the solution of equation~(\ref{equation_red_1}) is given by equation~(\ref{with_jum_adi}) and (\ref{without_jum_adi}). In Section~\ref{sec2}, we address inference for equation~(\ref{equation_red_1}), assuming that the jump distribution \( \nu \) and the parameters \( \lambda \) and \( \gamma \) are known. Appendix~C provides algorithms for estimating these parameters.

Our proposal is to consider an approach inspired by \cite{fang2022end}; however, under this framework, it is necessary to compute the conditional expectation of the approximation~(\ref{aproxh}) given the $\sigma$-algebra $\mathcal{F}(X_t)$ (we denote by $\mathcal{F}(\mathcal{H})$ the sigma-algebra generated by a family of random variables $\mathcal{H}$), as well as its associated likelihood function. As a first step, we compute this conditional expectation, and to evaluate the corresponding likelihood function, we estimate it through its characteristic function , as presented in \cite{Yang}.

Under the assumption that $v$ is a finite and symmetric measure on $\mathbb{Z}$, it follows that $E(z_i) = 0$ for $i = 1, \dots, N((t,t+\Delta t], \mathbb{Z})$.  Therefore, we have the following expectation:
\begin{equation}\label{expected_aprox}
    E\left(X_{t+\Delta t} \mid \mathcal{F}(X_t) \right) = X_t + f^{\Delta t}(X_t) \Delta t.
\end{equation}
Based on $\phi_{t,\Delta t|X_t}(u):= E(e^{iuX_{t+\Delta t}}|\mathcal{F}(X_t))$, Appendix~A, and Theorem~6 in \cite{Yang}, the \textit{approximated density function} of (\ref{conmil2}) given \( \mathcal{F}(X_t) \), which we denote by \( f_{t,\Delta t|X_t}^{M,h,a} \), is defined for fixed \( M \in \mathbb{N} \), \( h > 0 \), and \( a \in \mathbb{R} \) as:
\begin{equation}\label{fourierden}
    f_{t,\Delta t|X_t}^{M,h,a}(x) := \frac{1}{2\pi} \sum_{m=-M}^{M} e^{-ix(mh+ia)} \phi_{t,\Delta t|X_t}(mh+ia) h
\end{equation}

In Appendix~B, we present numerical studies using the Monte Carlo method to estimate the approximated density function in~(\ref{fourierden}), and compare this estimation with simulated samples. We observe that the approximated likelihood function shows strong agreement with the histogram of the simulated data.

Let $N \in \mathbb{N}$, $T > 0$, and \(\bar{t} = (t_1, \dots, t_N)\) be an increasing partition of the interval \([0, T]\), where \(t_1 = 0\) and \(t_N = T\). Define \(\Delta_{i+1} = t_{i+1} - t_i\) for \(i = 1, \dots, N-1\). Let \( z_i^{(s,t)} \) denote the \( i \)-th jump occurring in the interval \( (s,t] \), for \( i = 1, \dots, N((s,t], \mathbb{Z}) \). Since the random variables \( z_i^{(s,t)} \) are independent of the processes \( (W_t)_{t \geq 0} \) and \( (N((0,t], \mathbb{Z}))_{t \geq 0} \), and both of these processes have independent increments, it follows that the process \( \mathcal{X}(t_1,...,t_N) := (X_{t_i})_{i=1}^{N} \) forms a Markov chain with continuous state space. Therefore, using the approximated density function \(f_{t,\Delta t|X_t}^{M,h,a}\), the \textit{approximated likelihood function} for \(\mathcal{X}(t_1,...,t_N)\) is given by:
\begin{equation}\label{likelihood_aprox}
  f_{t_1,t_2,...,t_N}^{M,h,a}(f,g|(x_{t_1},...,x_{t_N})):= \Pi_{i=1}^{N-1} f_{t_i,\Delta t_i|x_{t_i}}^{M,h,a}(x_{t_{i+1}})
\end{equation}
for every realization \((x_{t_1}, \dots, x_{t_N})\) of the random vector \((X_{t_1}, \dots, X_{t_N})\). Given that an approximated likelihood function can be considered for the random vector $(X_{t_1}, \dots, X_{t_N})$ based on equation~(\ref{likelihood_aprox}), we adopt the two-step estimation method based on neural networks proposed by \cite{fang2022end}, which is described below.

Suppose we observe \(K\) realizations of equation~(\ref{conmil2}) evaluated at the time vector \(\bar{t}\), and denote the \(i\)-th realization by \(x^i\). We partition \(\bar{t}\) into \(R\) contiguous blocks, denoted by \((B_1, \dots, B_R)\), and let \(x_{j,t_k}^i\) represent the \(i\)-th realization restricted to time block \(B_j\), evaluated at time \(t_k \in B_j\).  Here, \(R\) corresponds to the number of batches used in the neural network training. We assume that \( |B_k| > 1 \), so that each time block contains at least two observation times, allowing us to define local dynamics within each block. In what follows, the notation \( \hat{f} \) and \( \hat{g} \) will be used to indicate that the loss functions, to be defined later in the paper, are evaluated on the approximations \( \hat{f} \) and \( \hat{g} \) to the drift and diffusion coefficients \( f \) and \( g \), respectively.

\textbf{Step 1:} In the first step, the drift coefficient $f$ is estimated using neural networks. The training objective is to minimize the mean squared error between the observed value and its conditional expectation given the previous state, as approximated by equation~(\ref{expected_aprox}). We define the lost function:\footnotesize
\begin{multline}\label{D1}
D_1(\hat{f},\hat{g}, B_k,j) := \left| \frac{1}{|B_k| - 1} \sum_{t_i \in B_k \setminus \{t_{\sum_{s=1}^k |B_s|}\}} \right. \\
\left. \left( x^{j}_{k,t_{i+1}} - x^{j}_{k,t_i} - \frac{\hat{f}(x^{j}_{k,t_i})}{1 + \Delta_{i+1} \hat{f}^2(x^{j}_{k,t_i})} \Delta_{i+1} \right)^2 \right|,\\ j=1,\dots,K,\quad k=1,\dots,R.
\end{multline}
\normalsize
\textbf{Step 2:} Given the estimated drift coefficient $f$, we propose to estimate the diffusion coefficient $g$ using neural networks. The loss function is defined as the approximated likelihood function given in equation~(\ref{likelihood_aprox}). Specifically,
\begin{multline}\label{loss_like}
D_2(\hat{f},\hat{g}, B_k,j) := - \sum_{t_i \in B_k \setminus \{t_{\sum_{s=1}^k |B_s|}\}} 
\log\left(f_{t_i,\Delta t_i|x_{k,t_{i+1}}^j}^{M,h,a}(x_{k,t_{i+1}}^j)\right), \\
j = 1, \dots, K,\quad k = 1, \dots, R.
\end{multline}
However, the original approach tends to yield poor estimates for the diffusion coefficient in scenarios where the solution to the SDE converges to a constant as \(t \to \infty\). This issue is illustrated in the following example. Consider the SDE:
\begin{equation}\label{EC0}
    dX(t) = a X(t)^3\,dt + b X(t)\,dW_t, \quad \text{where } a < 0, \; b \in \mathbb{R} \setminus \{0\}.
\end{equation}
As a consequence of the law of the iterated logarithm, $\lim_{t \to \infty} \frac{W_t}{t} = 0$ a.s., and based on this, it is easy to check that $\lim_{t \to \infty} X(t) = 0$ a.s.

We now present a case study in which equation~\eqref{EC0} is considered with parameters \(a = -0.25\) and \(b = 0.57\), and numerical simulations are performed using the approximation scheme defined in equation~\eqref{conmil2}. In this experiment, \(K = 10\) independent trajectories of the process are simulated over the interval \([0,5]\), using a uniform discretization with \(N = 1000\) time points. In each realization, the initial condition is \(X(0) = 1.5\), and the trajectories are generated using the scheme defined in equation~\eqref{conmil2}. 

\begin{figure}[ht]
    \centering
    \includegraphics[width=9cm,height=5cm]{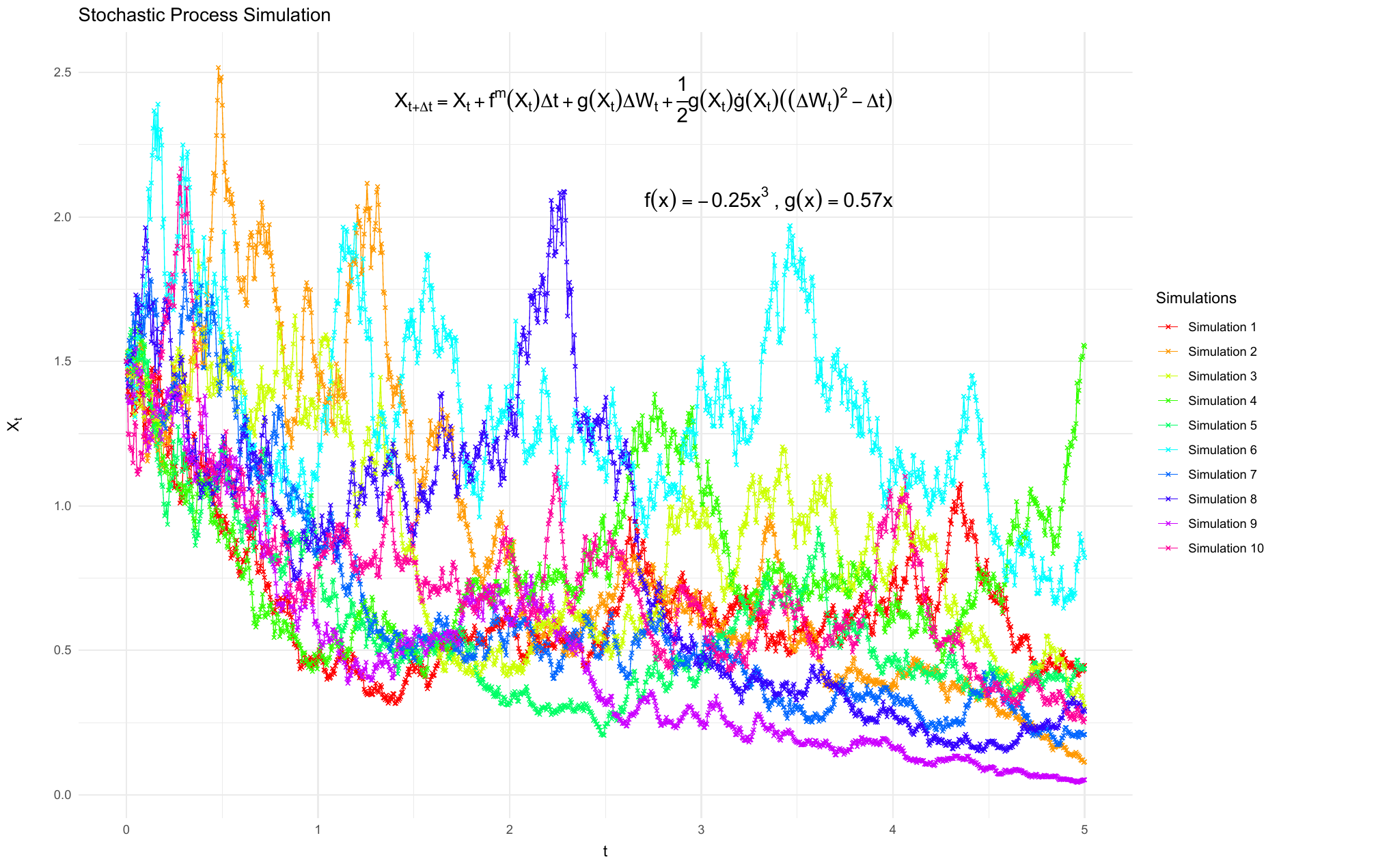}
    \caption{Sample trajectories of the process \(dX(t) = -0.25 X(t)^3\,dt + 0.57 X(t)\,dW_t\) with initial condition \(X(0) = 1.5\), simulated over the interval \([0,5]\) using the Tamed Milstein scheme defined in equation~(\ref{conmil2}). }
    \label{fig:sim_t}
\end{figure}

The neural networks used in this study were implemented and trained in the R programming environment using the \texttt{torch} library, which provides tools for building and optimizing deep learning models with GPU support. The simulated data are organized into consecutive batches of size 100, without shuffling (i.e., using \texttt{shuffle = FALSE}), in order to preserve the temporal structure of the trajectories. The resulting 10 simulated paths are shown in Figure~\ref{fig:sim_t}. The estimation procedure follows the two-step method described previously. In the first step, the drift coefficient \(f\) is estimated by minimizing the loss function \(D_1(\hat{f}, \hat{g}, B_{k},j)\); once this estimate is obtained, the second step consists in estimating the diffusion coefficient \(g\) by minimizing the loss function \(D_2(\hat{f}, \hat{g}, B_{k},j)\).

The function \(f\) is approximated by a neural network with a linear input layer, followed by four hidden layers with 32 neurons each, all using the exponential linear unit (ELU) activation function, and a linear output layer. The model is trained for 400 epochs using the Adam optimizer with a learning rate of 0.001, minimizing the loss function \(D_1(\hat{f}, \hat{g}, B_{k},j)\), evaluated on each batch \(B_k\) from each realization \(j\). Subsequently, the estimation of \(g\) is performed using a separate neural network with a linear input layer, followed by three hidden layers with 32 neurons each and ELU activations, and an output layer with Softplus activation to ensure the positivity of the estimated coefficient. This model is trained for 30 epochs using the Adam optimizer with the same learning rate, minimizing the loss function \(D_2(\hat{f}, \hat{g}, B_{k},j)\), evaluated over the same batches and realizations, using the previously obtained estimate of \(f\). For the computation of \(D_2(\hat{f}, \hat{g}, B_{k},j)\), we employed the approximation defined in equation~\eqref{fourierden} with parameters \(M = 200\), \(h = 0.05\), and \(a = 0\).

\begin{figure}[ht]
    \centering
    \includegraphics[width=9cm,height=4cm]{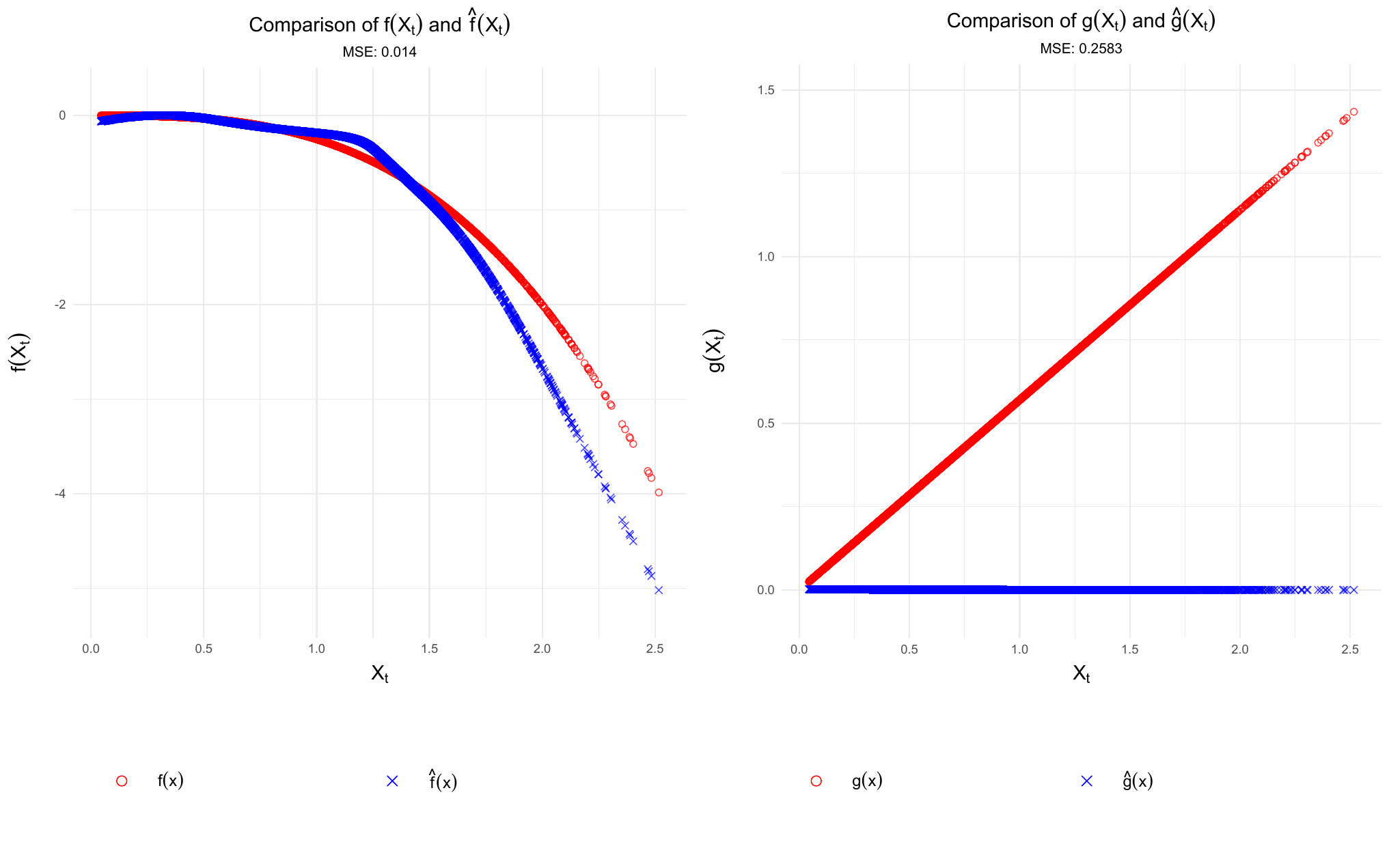}
    \caption{Estimation of the drift and diffusion coefficients \(f\) and \(g\) associated with the simulations in Figure~\ref{fig:sim_t}, using the two-step method described in \cite{fang2022end}. While the drift coefficient \(f\) is accurately recovered, the diffusion coefficient \(g\) is underestimated due to the asymptotic concentration of trajectories around zero.
}
    \label{fig:red_1}
\end{figure}

The results are presented in Figure~\ref{fig:red_1}. Since the solution of equation~\eqref{EC0} converges almost surely to zero. This behavior directly affects the likelihood-based estimation of the diffusion coefficient \(g\). In the second step of the two-stage procedure proposed in \cite{fang2022end}, \(g\) is estimated via a neural network by minimizing a loss function derived from transition density function. However, in regions where the sample paths converge and exhibit vanishing variability, the neural network tends to produce diffusion values close to zero, regardless of the true behavior of \(g\) outside those regions. This localized behavior biases the overall prediction of \(g\), resulting in a global diffusion estimate that is close to zero.

\begin{figure}[ht]
    \centering
    \includegraphics[width=9cm,height=4cm]{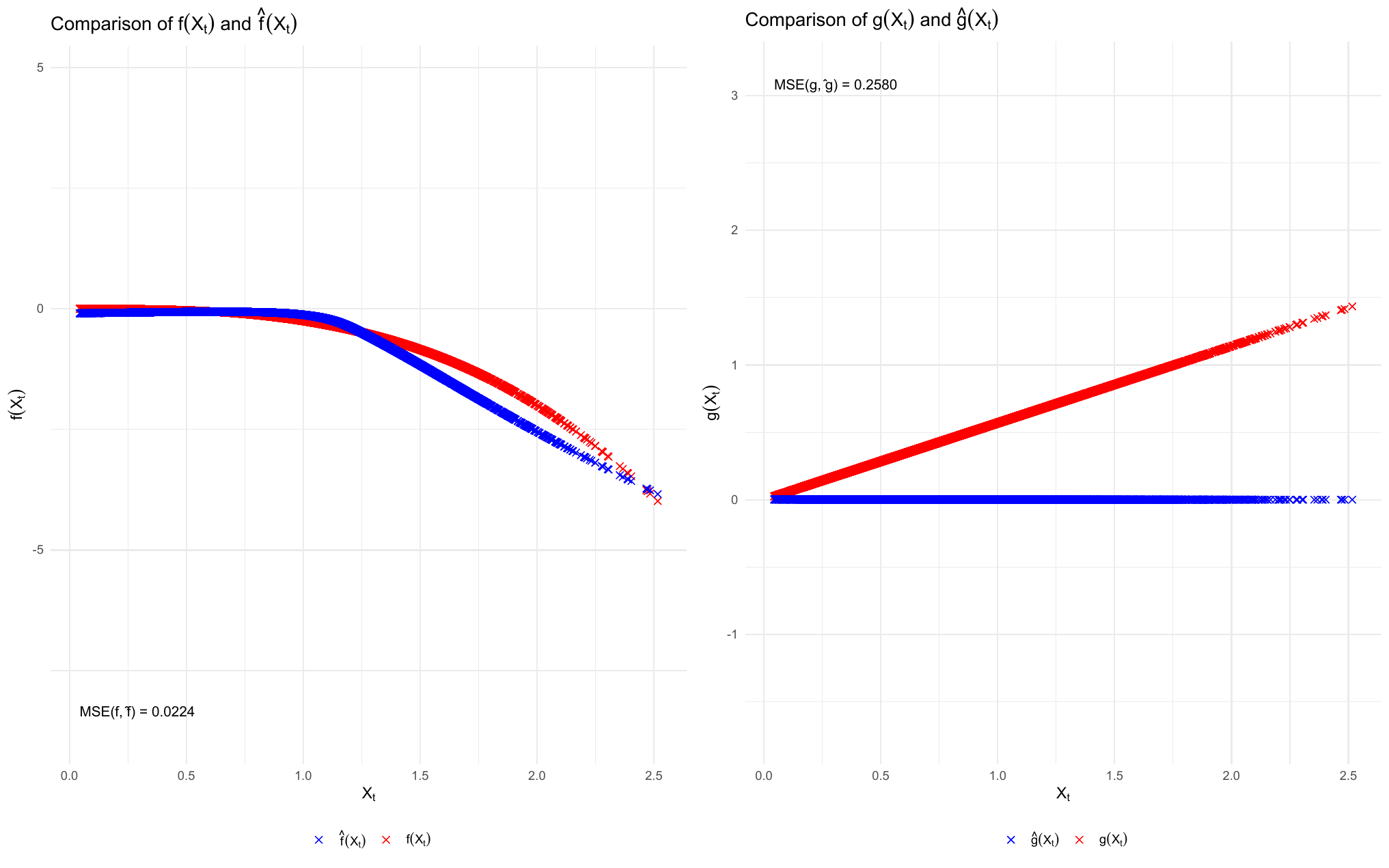}
    \caption{Estimation of the drift and diffusion coefficients \(f\) and \(g\) associated with the simulations in Figure~\ref{fig:sim_t}, using a single-step procedure based on the loss function \(D_2(\hat{f}, \hat{g}, B_{k},j)\) and the approximated likelihood described in equation~(\ref{likelihood_aprox}) with parameters \(M = 200\), \(h = 0.05\), and \(a = 0\). Both coefficients are trained jointly over 30 epochs.
}
    \label{fig:red_2}
\end{figure}

As shown in Figure~\ref{fig:red_1}, the method yields a coherent approximation of the drift coefficient \(f\), but it fails to recover a meaningful estimate of the diffusion coefficient \(g\).
Furthermore, Figure~\ref{fig:red_2} shows that under the same modeling setup, similar results can be obtained employing a single-step estimation procedure in which both \(f\) and \(g\) are trained simultaneously using the loss function \(D_2(\hat{f}, \hat{g}, B_{k},j)\). In this case, the number of training epochs is set to 30 for the joint optimization of \(f\) and \(g\), and the resulting estimates are comparable to those obtained with the two-step procedure previously described. Although the mean squared error (MSE) for the drift coefficient \(f\) is lower under the two-step training strategy, the single-step method based on \(D_2(\hat{f}, \hat{g}, B_{k},j)\) appears to yield a better global structural approximation for \(f\). That is, the approximated likelihood function \(f_{t_1,t_2,\dots,t_N}^{M,h,a}(f,g|(x_{t_1}, \dots, x_{t_N}))\) provides an effective fit for the estimation of the drift coefficient \(f\).

In Section~\ref{sec:Algorithm}, we provide an algorithm for the estimation of the drift and diffusion coefficients, introducing a random selective training strategy designed to prevent overfitting of the diffusion coefficient \(g\) near regions where the trajectories converge.

\subsection{Algorithm}\label{sec:Algorithm}

In this section, we provide an algorithm for estimating the drift and diffusion coefficients. Our approach relies on a selective random training scheme aimed at preventing overfitting of the diffusion coefficient near regions of convergence. We assume that the architecture of the neural networks associated with the coefficients $f$ and $g$ is as previously described. The algorithm consists of three phases, which we detail below:

\textbf{Phase 1:} The first phase takes advantage of the fact that the approximate likelihood function given by (\ref{fourierden}), through the loss function (\ref{loss_like}), allows us to identify the structure of the drift coefficient $f$ by simultaneously training the neural networks associated with $f$ and $g$. This training is carried out over a fixed number of epochs, denoted by $epoch_0$. Once this training step is completed, we reset the neural network associated with the diffusion coefficient $g$.

\textbf{Phase 2:} We now consider the additional assumption that the jump sizes \( z_i \) satisfy \( E(z_i^2) < \infty \). In this phase, the goal is to minimize loss functions designed to reduce the conditional mean and variance of the approximation (\ref{conmil2}) given $\mathcal{F}(X_t)$, focusing on branches associated with higher loss values. The selection is performed randomly in order to avoid overfitting the neural networks to a specific subset of branches.

For instance, near potential convergence regions, the conditional variance of (\ref{conmil2}) given $\mathcal{F}(X_t)$ is expected to be small. As a result, the probability of training on branches close to those regions is reduced. Since we define loss functions based on the conditional variance of the approximation (\ref{conmil2}) given $\mathcal{F}(X_t)$, this quantity is computed as follows:
In the case of a diffusion without jumps,  we have that:
\begin{multline}\label{sec_moment}
E\left(\left(X_{t+\Delta t} - E(X_{t+\Delta t} \mid \mathcal{F}(X_t))\right)^2 \mid \mathcal{F}(X_t)\right) \\
= g^2(X_t)\Delta t + \frac{1}{2}\left(g(X_t)g^{'}(X_t)\Delta t\right)^2.
\end{multline}
And, for a jump-diffusion process, define
\begin{align*}
    M_1 &:= g(X_t) \Delta W_t + \frac{1}{2} g(X_t) g'(X_t) \left((\Delta W_t)^2 - \Delta t\right) \\
    &\quad + \sum_{i=1}^{N((t,t+\Delta t],\mathbb{Z})} \left(g(X_t + \gamma z_i^{(t,t+\Delta t)}) - g(X_t)\right)(W_{t+\Delta t} - W_{\tau_i}), \\
    M_2 &:= \gamma \sum_{i=1}^{N((t,t+\Delta t],\mathbb{Z})} z_i^{(t,t+\Delta t)}.
\end{align*}
It is straightforward to verify that \( E(M_1 M_2 \mid \mathcal{F}(X_t)) = 0 \) and \( E(M_2^2 \mid \mathcal{F}(X_t)) = \gamma^2 \mu_2 \lambda \Delta t \). Thus,\scriptsize
\begin{multline}\label{secondjump}
E\left(\left(X_{t+\Delta t} - E(X_{t+\Delta t} \mid \mathcal{F}(X_t))\right)^2 \mid \mathcal{F}(X_t)\right) = E(M_1^2 \mid \mathcal{F}(X_t)) + E(M_2^2 \mid \mathcal{F}(X_t)) \\
= g^2(X_t)\Delta t + \frac{1}{2}\left(g(X_t)g'(X_t)\Delta t\right)^2 \\
+ 2g(X_t) E\left(\sum_{i=1}^{N((t,t+\Delta t],\mathbb{Z})} \left(g(X_t + \gamma z_i^{(t,t+\Delta t)}) - g(X_t)\right)(t + \Delta t - \tau_i)\right) \\
+ E\Bigg(\sum_{i=1}^{N((t,t+\Delta t],\mathbb{Z})}\sum_{j=1}^{N((t,t+\Delta t],\mathbb{Z})} \left(g(X_t + \gamma z_i^{(t,t+\Delta t)}) - g(X_t)\right) \\
\cdot \left(g(X_t + \gamma z_j^{(t,t+\Delta t)}) - g(X_t)\right)(t + \Delta t - \tau_{i \vee j}) \Bigg) + \gamma^2 \mu_2 \lambda \Delta t > 0.
\end{multline}\normalsize
When the parameters \( \lambda \neq 0 \) and \( \gamma \neq 0 \), the conditional second moment given in equation~(\ref{secondjump}) can be estimated using the law of large numbers.

In this phase, the algorithm employs a selective alternating random training procedure. The first step consists of training the neural network associated with the drift coefficient $f$ in order to minimize the conditional first moment of $X_{t+\Delta t}$ given $\mathcal{F}(X_t)$. Once this training step is completed, and given the current estimate of the drift coefficient $f$, we aim to reduce the conditional variance of $X_{t+\Delta t}$ given $\mathcal{F}(X_t)$ by training the neural network associated with the diffusion coefficient $g$. We repeat this procedure for a fixed number of epochs, denoted by $epoch_1$. 

Next, we describe Phase 2 of the algorithm in more detail. We consider a random selection mechanism as follows. Let \(U_{1,j}\), for \(j=1,\dots,R\), be independent uniform random variables on the set \(\{1,\dots,K\}\). We denote by \( u_{1,j} \) the realization of the random variable \( U_{1,j} \) and we define the loss functions:
\begin{multline}
L_{1,j} := L_1(\hat{f}, \hat{g}, B_j,u_{1,j}) := \left| \frac{1}{|B_j| - 1} \sum_{t_i \in B_j \setminus \{t_{\sum_{s=1}^j |B_s|}\}} \right. \\
\left. \left( x_{j,t_{i+1}}^{u_{1,j}} - x_{j,t_i}^{u_{1,j}} - \frac{\hat{f}(x_{j,t_i}^{u_{1,j}})}{1 + \Delta_{i+1} \hat{f}^2(x_{j,t_i}^{u_{1,j}})} \Delta_{i+1} \right) \right|, \quad j = 1, \dots, R.
\end{multline}
We define \(w_{1,j} := \frac{L_{1,j}}{\sum_{i=1}^R L_{1,i}}\), and select \(R_1\) indices without replacement from the index set \([R] := \{1,\dots,R\}\) with weights \(\{w_{1,1}, \dots, w_{1,R}\}\), denoted \(i_{1,1}, \dots, i_{1,R_1}\). We then train the neural network corresponding to the function \(f\) over the batches \( \{ x_{i_{1,1},t_i}^{u_{1,i_{1,1}}} \mid t_i \in B_{i_{1,1}} \} \),...,\( \{ x_{i_{1,R_1},t_i}^{u_{1,i_{1,R_1}}} \mid t_i \in B_{i_{1,R_1}} \} \) using the loss function \(L_1 (\hat{f}, \hat{g}, B_{i_{1,k}},u_{1,i_{1,k}})\), $k=1,...,R_1$.

Similarly, let \(U_{2,j}\), be independent uniform random variables on the set \(\{1,\dots,K\}\). We denote by \( u_{2,j} \) the realization of the random variable \( U_{2,j} \) and we define the loss functions:
\scriptsize
\begin{multline}
L_2(\hat{f}, \hat{g}, B_j,u_{2,j}) := \frac{1}{|B_j| - 1} \sum_{t_i \in B_j \setminus \{t_{\sum_{s=1}^j |B_s|}\}} \Bigg( \Bigg( x_{j,t_{i+1}}^{u_{2,j}}  - x_{j,t_{i}}^{u_{2,j}}\\  
- \frac{\hat{f}(x_{j,t_{i}}^{u_{2,j}} )}{1 + \Delta_{i+1} \hat{f}^2(x_{j,t_{i}}^{u_{2,j}} )} \Delta_{i+1} \Bigg)^2 
- E\Bigg(\Bigg(X_{t+\Delta t} - E(X_{t+\Delta t} \mid X_t=x_{j,t_{i}}^{u_{2,j}})\Bigg)^2 \mid X_t=x_{j,t_{i}}^{u_{2,j}}\Bigg) \Bigg)^2,\\ 
\phantom{a} j = 1, \dots, R.
\end{multline}
\normalsize
 Let $L_{2,j} := L_2(\hat{f}, \hat{g}, B_j,u_{2,j})$,  \(w_{2,j} := \frac{L_{2,j}}{\sum_{i=1}^R L_{2,i}}\), and we select \(R_2\) indices without replacement from the index set \([R] := \{1,\dots,R\}\) with weights \(\{w_{2,1}, \dots, w_{2,R}\}\), denoted \(i_{2,1}, \dots, i_{2,R_2}\). We then train the neural network corresponding to the function \(g\) on the dataset \( \{ x_{i_{2,1},t_i}^{u_{2,i_{2,1}}} \mid t_i \in B_{i_{2,1}} \} \),...,\( \{ x_{i_{2,R_2},t_i}^{u_{2,i_{2,R_2}}} \mid t_i \in B_{i_{2,R_2}} \} \) using the loss function \(L_2 (\hat{f}, \hat{g}, B_{i_{2,k}},u_{2,i_{2,k}})\), $k=1,...,R_2$.

\begin{figure}[ht]
\centering
\includegraphics[width=9cm,height=4cm]{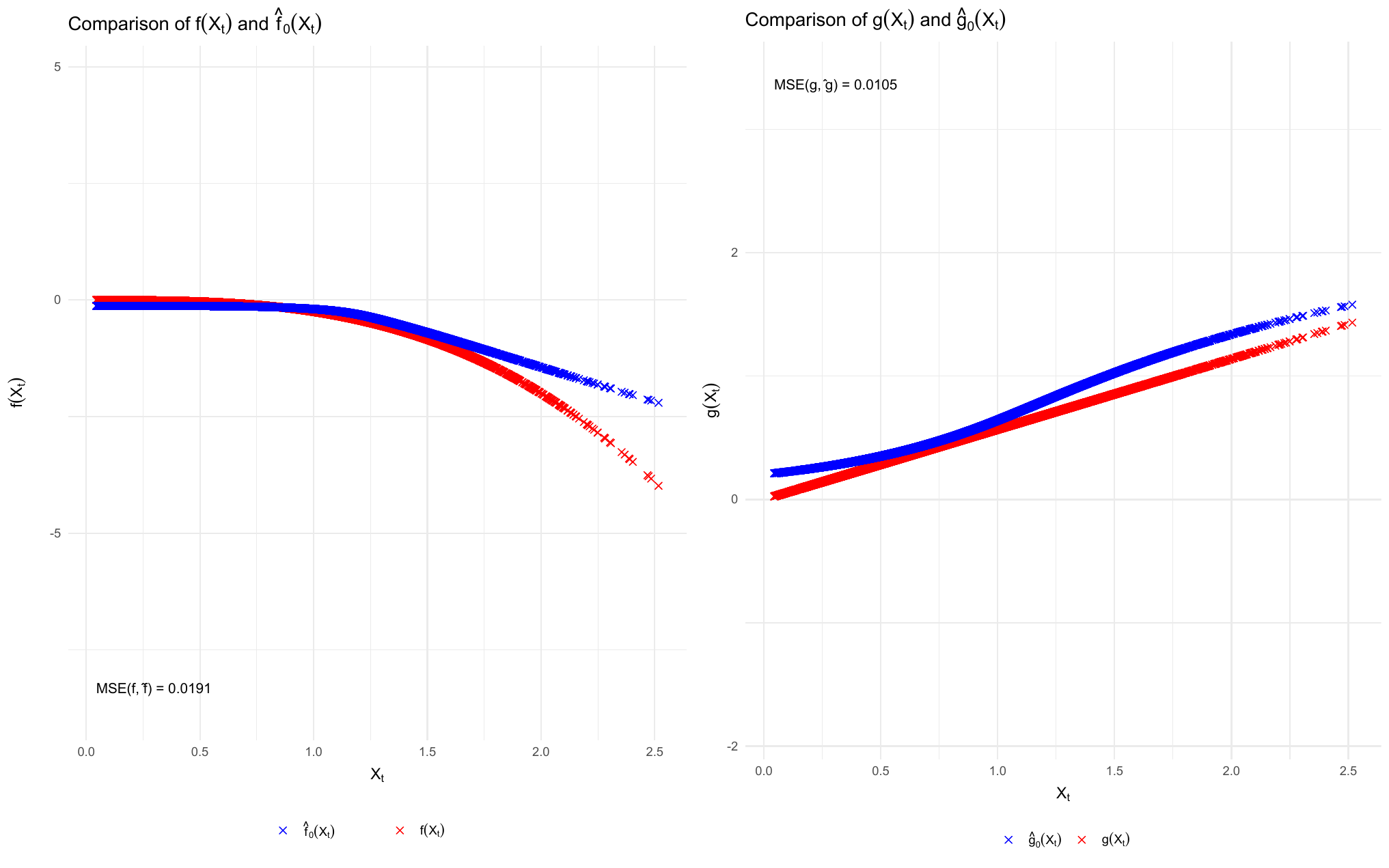}
\par\smallskip
\textit{Panel A: Phase 2 of the estimation algorithm.}
\par\medskip
\includegraphics[width=9cm,height=4cm]{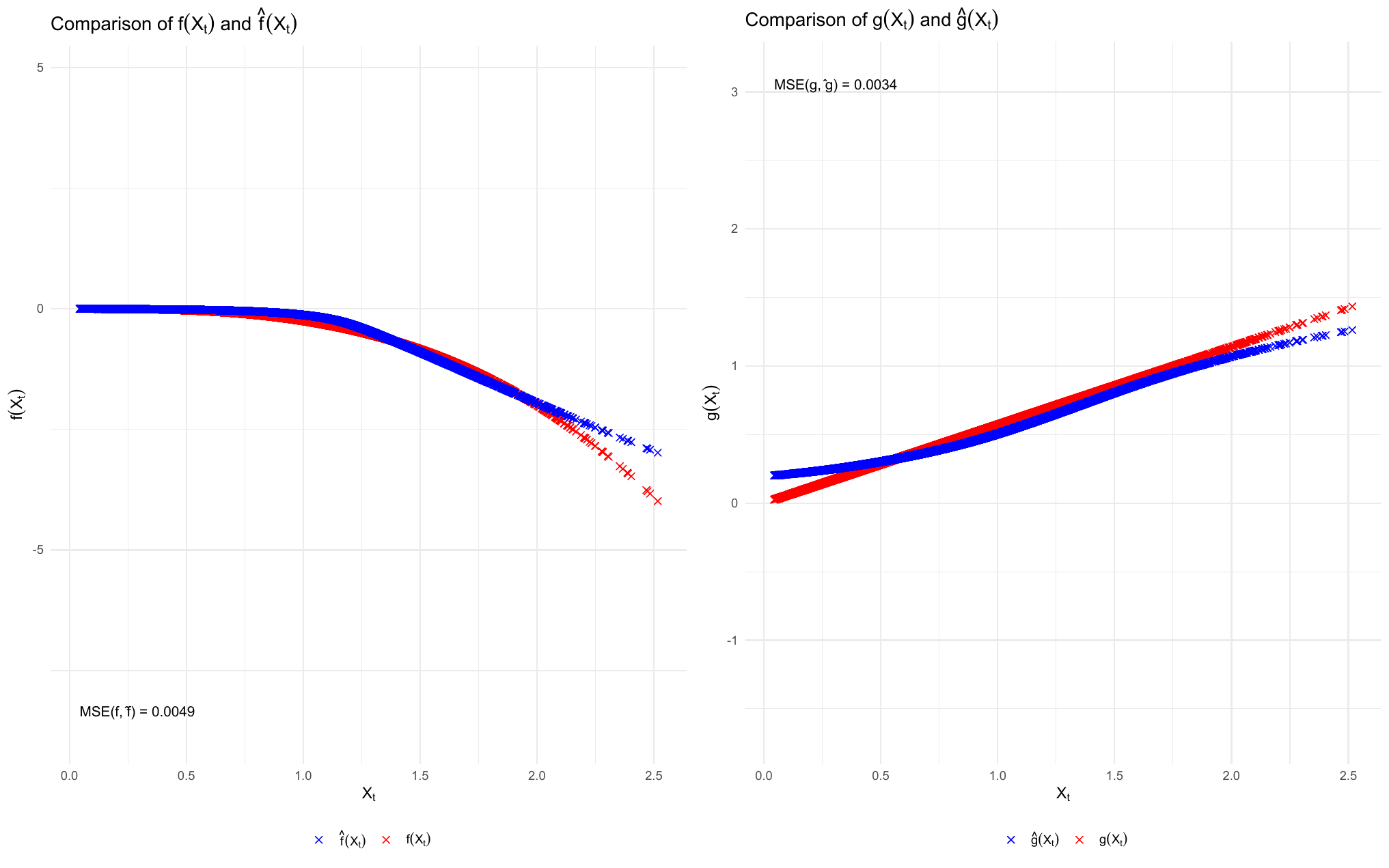}
\par\smallskip
\textit{Panel B: Phase 3 of the estimation algorithm.}
\caption{\label{fig:comparacion_likelihood}
Estimation of the drift and diffusion coefficients \(f\) and \(g\) based on the simulations shown in Figure~\ref{fig:sim_t}. Panel A shows the estimates \(\hat{f}_0\) and \(\hat{g}_0\) obtained in Phase 2, with \(epoch_1 = 100\), \(R_1 = 8\), and \(R_2 = 2\), yielding MSEs of 0.0191 and 0.0105 for the drift and diffusion coefficients, respectively. Panel B shows the estimates obtained in Phase 3, with \(epoch_1 = 400\), \(train_f = 4\), \(R_3 = 4\), and \(R_4 = 4\), yielding improved MSEs of 0.0049 and 0.0034.}
\end{figure}

This procedure allows us to obtain estimators for the coefficients \(f\) and \(g\), which we denote by \(\hat{f}_0\) and \(\hat{g}_0\). In Figure~\ref{fig:comparacion_likelihood}, Panel A, we present the training corresponding to this phase, applied to the simulated sample shown in Figure~\ref{fig:sim_t}. In this experiment, we used \(epoch_1 = 100\), \(R_1 = 8\) (equivalent to 80\%), and \(R_2 = 2\) (equivalent to 20\%). The mean squared error (MSE) for the neural network estimating the drift coefficient was 0.0191, while the MSE for the neural network estimating the diffusion coefficient was 0.0105.

Given the estimators \(\hat{f}_0\) and \(\hat{g}_0\), our objective is to obtain more accurate approximations. To this end, we adopt a strategy based on training the networks using standardized data sets. It is important to note that selection criteria based solely on the expected value or second moment may bias the training toward samples with higher variance. Therefore, we perform the selection based on standardized variance to ensure a balanced representation across the training set. 

Hence, using the conditional expectation defined in \eqref{secondjump}, we define:
\begin{equation}\label{Ysaltos}
    Y_{t,\Delta t}^{\gamma} := \frac{X_{t+\Delta t} - (X_t + f^{\Delta t}(X_t)\Delta t)}{\sqrt{E\left(\left(X_{t+\Delta t} - E(X_{t+\Delta t} \mid \mathcal{F}(X_t))\right)^2 \mid \mathcal{F}(X_t)\right)}}.
\end{equation}
and when $\gamma=0$
\begin{equation}\label{Ysin_saltos}
    Y_{t,\Delta t}^{\gamma} := 
    \begin{cases} 
        X_{t+\Delta t} - (X_t + f^{\Delta t}(X_t)\Delta t), & \text{if } g(X_t) = 0, \\
        \dfrac{X_{t+\Delta t} - (X_t + f^{\Delta t}(X_t)\Delta t)}
        {\sqrt{g^2(X_t)\Delta t + \frac{1}{2} \left(g(X_t)g^{'}(X_t)\Delta t\right)^2}}, 
        & \text{if } g(X_t) \neq 0.
    \end{cases}
\end{equation}
Then, by (\ref{expected_aprox}) and (\ref{sec_moment}) we have that $E(Y_{t, \Delta t}^{\gamma}\mid \mathcal{F}(X_t)) = 0$ and $\text{Var}(Y_{t, \Delta t}^{\gamma}\mid \mathcal{F}(X_t)) = 1$. We now proceed to describe Phase 3 of the algorithm. In this phase, the algorithm continues with the alternating training strategy for the neural networks associated with the drift and diffusion coefficients. The first step consists in training the neural network estimating the drift coefficient \(f\) by minimizing the conditional expectation and the conditional variance of \(X_{t+\Delta t}\) given \(\mathcal{F}(X_t)\). Once this step is completed, the neural network corresponding to the diffusion coefficient \(g\) is trained with the aim of minimizing the conditional variance of \(X_{t+\Delta t}\) given \(\mathcal{F}(X_t)\). The main difference with respect to Phase 2 is that the selection of training samples is now based on the standardized random variables defined in equation~(\ref{Ysin_saltos}), with the goal of constructing a more homogeneous training set by standardizing the data used in the selection procedure.

\textbf{Phase 3:}  We consider the following training mechanism: For each \(B_j\), \(j=1,\dots,R\), we define $B^{-j}:=B_j \setminus \{t_{\sum_{s=1}^{j}|B_s|}\}$ and $B^{i}_{j,\hat{g}} := \{t_k \in B^{-j}\phantom{a}|\phantom{a}  \hat{g}(x_{j,t_k}^{i}) \neq 0\}, \phantom{a}i=1,...,K, \phantom{a}j=1,...,R$. We denote by \( y^{\gamma,i}_{t_k,\Delta t_k} \) the realization of the random variable \( Y^{\gamma}_{t_k,\Delta t_k} \), for \( k = 1, \dots, N-1 \), based on the trajectory \( x^i \) according to equations~(\ref{Ysin_saltos}) and~(\ref{Ysaltos}). Let \(U_{3,j}\), be independent uniform random variables on the set \(\{1,\dots,K\}\) and we denote by \( u_{3,j} \) the realization of the random variable \( U_{3,j} \). Based on equations~(\ref{Ysaltos}) and~(\ref{Ysin_saltos}), for \( j = 1, \dots, R \), we define \( H_j \) as follows:
\begin{multline*}
H_{j} := H(\hat{f},\hat{g},B_j,u_{3,j}) := 
\frac{1}{|B_{j,\hat{g}}^{u_{3,j}}|} \sum_{t_i \in B_{j,\hat{g}}^{u_{3,j}}} (y^{\gamma,u_{3,j}}_{t_i,\Delta t_i})^2 \cdot \mathbf{1}_{\{|B_{j,\hat{g}}^{u_{3,j}}| \neq 0\}} \\
+ \frac{1}{|B^{-j} \cap (B^{u_{3,j}}_{j,\hat{g}})^c|} \sum_{t_i \in B^{-j} \cap (B^{u_{3,j}}_{j,\hat{g}})^c} (y^{\gamma,u_{3,j}}_{t_i,\Delta t_i})^2 \cdot \mathbf{1}_{\{|B^{-j} \cap (B^{u_{3,j}}_{j,\hat{g}})^c| \neq 0\}}.
\end{multline*}
Given that \(E\left((Y_{t, \Delta t}^{\gamma})^2 \mid \mathcal{F}(X_t)\right) = 1\), the law of large numbers implies that \(H_{j} \approx 1\) and \(\frac{1}{H_{j}} \approx 1\). This phase proceeds in two steps. First, the neural network associated with the drift coefficient is trained to minimize both the conditional expectation and conditional variance of \(X_{t+\Delta t}\) given \(\mathcal{F}(X_t)\), using a selective random training procedure that favors subsets with larger values of \(H_{j}\). Once the drift coefficient \(f\) has been estimated, the second step consists in training the neural network associated with the diffusion coefficient \(g\) to minimize the expected conditional variance of \(X_{t+\Delta t}\) given \(\mathcal{F}(X_t)\), this time favoring subsets with smaller values of \(H_{j}\).

We will repeat the following two steps for a number of \(\text{epoch}_2\) iterations. We will repeat the next step \(\text{train}_f\) times.  For \(j = 1,\dots,R\), define \(w_{3,j} = \frac{H_{j}}{\sum_{i=1}^R H_{i}}\), and take a random sample without replacement of size \(R_3\) from the index set \([R] := \{1,\dots,R\}\), with weights \(\{w_{3,1},\dots,w_{3,R}\}\), denoted \(i_{3,1},\dots,i_{3,R_3}\). Then we train the neural network associated with the function \(f\) using the data sets 
\(\{ x_{i_{3,1},t_i}^{u_{3,i_{3,1}}} \mid t_i \in B_{i_{3,1}} \}, \ldots, \{ x_{i_{3,R_3},t_i}^{u_{3,i_{3,R_3}}} \mid t_i \in B_{i_{3,R_3}} \}\), 
and the loss function 
\[
L_3(\hat{f}, \hat{g}, B_{j}, u_{3,j}) + L_4(\hat{f}, \hat{g}, B_{j}, u_{3,j}), \quad j = 1, \ldots, R_3,
\]
where the loss functions \(L_3\) and \(L_4\) are defined as follows:
\footnotesize
\begin{multline*}
L_3(\hat{f},\hat{g},B_j,u_{3,j}) :=\\
\frac{1}{|B_{j,\hat{g}}^{u_{3,j}}|} 
\sum_{t_i \in B_{j,\hat{g}}^{u_{3,j}}} 
\Bigg(x_{j,t_{i+1}}^{u_{3,j}} - x_{j,t_{i}}^{u_{3,j}} 
- \frac{\hat{f}(x_{j,t_{i}}^{u_{3,j}})}{1 + \Delta_{i+1} \hat{f}^2(x_{j,t_{i}}^{u_{3,j}})} \Delta_{i+1} \Bigg)^2 
\cdot \mathbf{1}_{\{|B_{j,\hat{g}}^{u_{3,j}}| \neq 0\}}\\
+\frac{1}{|B^{-j} \cap (B^{u_{3,j}}_{j,\hat{g}})^c|} 
\sum_{t_i \in B^{-j} \cap (B^{u_{3,j}}_{j,\hat{g}})^c} 
\Bigg(x_{j,t_{i+1}}^{u_{3,j}} - x_{j,t_{i}}^{u_{3,j}} 
- \frac{\hat{f}(x_{j,t_{i}}^{u_{3,j}})}{1 + \Delta_{i+1} \hat{f}^2(x_{j,t_{i}}^{u_{3,j}})} \Delta_{i+1} \Bigg)^2 \\
\cdot \mathbf{1}_{\{|B^{-j} \cap (B^{u_{3,j}}_{j,\hat{g}})^c| \neq 0\}}, 
\end{multline*}
and
\begin{multline*}
L_4(\hat{f},\hat{g},B_j,u_{3,j}) := \\
\Bigg[
\frac{1}{|B_{j,\hat{g}}^{u_{3,j}}|} \sum_{t_i \in B_{j,\hat{g}}^{u_{3,j}}}
\Bigg(
\Bigg(x_{j,t_{i+1}}^{u_{3,j}} - x_{j,t_i}^{u_{3,j}} - \frac{\hat{f}(x_{j,t_i}^{u_{3,j}})}{1 + \Delta_{i+1} \hat{f}^2(x_{j,t_i}^{u_{3,j}})} \Delta_{i+1} \Bigg)^2 \\
- E\Bigg( \Bigg( X_{t+\Delta t} - \mathbb{E}(X_{t+\Delta t} \mid X_t = x_{j,t_i}^{u_{3,j}}) \Bigg)^2 \,\big|\, X_t = x_{j,t_i}^{u_{3,j}} \Bigg)
\Bigg)^2
\Bigg] \cdot \mathbf{1}_{\{|B_{j,\hat{g}}^{u_{3,j}}| \neq 0\}} \\
+ \Bigg[
\frac{1}{|B^{-j} \cap (B^{u_{3,j}}_{j,\hat{g}})^c|} \sum_{t_i \in B^{-j} \cap (B^{u_{3,j}}_{j,\hat{g}})^c}
\Bigg(\Bigg(x_{j,t_{i+1}}^{u_{3,j}} - x_{j,t_i}^{u_{3,j}} - \frac{\hat{f}(x_{j,t_i}^{u_{3,j}})}{1 + \Delta_{i+1} \hat{f}^2(x_{j,t_i}^{u_{3,j}})} \Delta_{i+1} \Bigg)^2\\
- E\Bigg( \Bigg( X_{t+\Delta t} - \mathbb{E}(X_{t+\Delta t} \mid X_t = x_{j,t_i}^{u_{3,j}}) \Bigg)^2 \,\big|\, X_t = x_{j,t_i}^{u_{3,j}} \Bigg)
\Bigg)^2
\Bigg] \cdot \mathbf{1}_{\{|B^{-j} \cap (B^{u_{3,j}}_{j,\hat{g}})^c| \neq 0\}}
\end{multline*}
\normalsize
Let \(U_{4,j}\), be independent uniform random variables on the set \(\{1,\dots,K\}\) and we denote by \( u_{4,j} \) the realization of the random variable \( U_{4,j} \). For \(j = 1,\dots,R\), define \(w_{4,j} = \frac{1/H_{j}}{\sum_{i=1}^R 1/H_{i}}\), and select \(R_4\) indices without replacement from the index set \([R] := \{1,\dots,R\}\) with weights \(\{w_{4,1},\dots,w_{4,R}\}\), denoted \(i_{4,1},\dots,i_{4,R_4}\). We then train the neural network associated to the function \(g\) on the dataset \( \{ x_{i_{4,1},t_i}^{u_{4,i_{4,1}}} \mid t_i \in B_{i_{4,1}} \} \),...,\( \{ x_{i_{4,R_4},t_i}^{u_{4,i_{4,R_4}}} \mid t_i \in B_{i_{4,R_4}} \} \) using the loss function \(L_2 (\hat{f}, \hat{g}, B_{i_{4,k}},u_{4,i_{4,k}})\), $k=1,...,R_4$.

In Figure~\ref{fig:comparacion_likelihood}, Panel B, we present the training corresponding to this phase, applied to the simulated sample shown in Figure~\ref{fig:sim_t}. In this experiment, we used \(epoch_1 = 400\), \(train_f = 4\), \(R_3 = 4\) (equivalent to 40\%), and \(R_4 = 4\) (equivalent to 40\%). The MSE for the neural network estimating the drift coefficient was 0.0049, while the MSE for the neural network estimating the diffusion coefficient was 0.0034. The neural network approximation using the proposed algorithm yields a MSE of order \(10^{-3}\) for the estimation of the drift and diffusion coefficients \(f\) and \(g\).
% In Section~\ref{Numerical_Studies_AJ}, we present several numerical examples in which the methodology described in this section is applied to simulated data for different functions \( f \) and \( g \) that satisfy assumptions \textbf{(A2)}–\textbf{(A5)}, under different values of the parameters \( \lambda \) and \( \gamma \), and with different distributions associated with the jump magnitudes \( z_i \).
\subsection{Numerical Studies}\label{Numerical_Studies_AJ} 

In this section, we present several numerical studies based on simulated data to evaluate the methodology described in Section~\ref{sec2} for estimating the drift and diffusion coefficients \(f\) and \(g\). 

In the following experiments, \(K = 10\) independent trajectories of the process are simulated over the interval \([0,5]\), using a uniform discretization with \(N = 1000\) time points. In each realization, the initial condition is \(X(0) = 1.5\), and the trajectories are generated using the scheme defined in equation~\eqref{conmil2}. The functions \(f\) and \(g\) satisfy assumptions \textbf{(A2)}–\textbf{(A5)}. The neural networks used to estimate the coefficients \(f\) and \(g\) are the same as those described in Section~\ref{sec2}. The training parameters \(R_1, R_2, R_3, R_4\) and \(\text{train}_f\) are also kept unchanged, as is the numbers of training epochs \(\text{epoch}_0\), \(\text{epoch}_1\) and \(\text{epoch}_2\).

Table~\ref{table:experiments} presents numerical examples in which the parameters associated with the jump component are set to zero. In contrast, Table~\ref{table:experiments2} shows numerical examples with jumps, where the distributions of the jump sizes are symmetric about zero and have finite second moments. For these experiments, we approximate equation~\eqref{secondjump} using simulations and the law of large numbers with a sample of 400 trajectories simulated over the interval \([t, t+\Delta)\). To estimate the approximated density function \(f_{t,\Delta t|X_t}^{M,h,a}(x)\), we use Monte Carlo integration: we simulate 200 realizations of the random variables \(N((t, t+\Delta], Z)\), \((\tau_1, \ldots, \tau_{N((t, t+\Delta], Z)})\), and \((z_1, \ldots, z_{N((t, t+\Delta], Z)})\). The density estimation is carried out using equation~\eqref{fourierden}, with parameters \(M = 200\), \(h = 0.05\), and \(a = 0\) (see Appendix~B). In these experiments, all training parameters are kept fixed, except for the number of training epochs, which is set to \(\text{epoch}_0 = 10\). In both settings, and under various scenarios, the methodology proposed in Section~\ref{sec2} for estimating the drift and diffusion coefficients \(f\) and \(g\) yields satisfactory results, with the maximum reported MSE between the true functions and the network estimations being on the order of \(10^{-2}\).

\tiny
\begin{table}
\caption{\label{table:experiments}
Numerical studies based on simulated data for the estimation of the coefficients \(f\) and \(g\) using the methodology proposed in Section~\ref{sec2}, in scenarios where either \(\lambda = 0\) or \(\gamma = 0\).}
\begin{ruledtabular}
\begin{tabular}{cccc}
\makecell{Drift\\coefficient $f$} & \makecell{Diffusion\\coefficient $g$} & \multicolumn{1}{c}{\textbf{Error $L_2f$}} & \multicolumn{1}{c}{\textbf{Error $L_2g$}} \\
\hline
 $-0.25X_t^3$ & $0.57X_t$ & 0.00492 & 0.00347 \\
$0.15(X_t - X_t^5)$ & $0.32\sin(X_t)$ & 0.00292 & 0.00013 \\
 $1 - X_t$ & $1$ & 0.01600 & 0.00009 \\
 $\sin(X_t)$ & $1$ & 0.03597 & 0.00033 \\
\end{tabular}
\end{ruledtabular}
\end{table}

{\fontsize{5}{6}\selectfont
\begin{table}
\caption{\label{table:experiments2}
Numerical studies based on simulated data for the estimation of the coefficients \(f\) and \(g\) using the methodology proposed in Section~\ref{sec2}, in scenarios where either \(\lambda \neq 0\) and \(\gamma \neq 0\).}
\begin{ruledtabular}
\begin{tabular}{cccccc}
\makecell{Parameters\\$(\gamma,\lambda)$} & \makecell{Jump\\$z_i$} & \makecell{Drift\\$f(X_t)$} & \makecell{Diffusion\\$g(X_t)$} & \makecell{$L_2f$} & \makecell{$L_2g$} \\
\hline
(0.8, 1.2)   & $U(-0.1,0.1)$     & $1-X_t $           & $0.31X_t$                  & 0.00401 & 0.00046 \\
(0.31, 1.7)  & $N(0,0.12)$       & $0.28(X_t-X_t^3) $       & $1$        & 0.05564 & 0.00094 \\
(1.47, 0.5)   & Laplace(0,0.1)    & $\cos(X_t)$            & $1$                        & 0.00675 & 0.00008 \\
\end{tabular}
\end{ruledtabular}
\end{table}
}
\normalsize
For the selection of the training parameters \((R_1, R_2, R_3, R_4,\text{trainf}_f)\), one may consider a validation set and perform training over a predefined grid of candidate values. The optimal parameters are then selected based on performance over the validation set. In the case of simulated data, the selection criterion can be the mean squared error (MSE) between the true coefficients and the estimates produced by the neural networks. In the case of real data, where the true drift and diffusion coefficients \(f\) and \(g\) are unknown, one can simulate the stochastic differential equation~(\ref{conmil2}) using the estimated coefficients, and compute the MSE between the sample means of the simulated trajectories and the observed validation data points.

\section{Conclusions}\label{Discussion}

In Section~\ref{sec2}, we proposed a methodology for the estimation of the drift and diffusion coefficients \(f\) and \(g\) in the stochastic differential equation~(\ref{equation_red_1}). This methodology is based on the Tamed Milstein approximation scheme introduced in equation~(\ref{conmil2}), and relies on neural networks to perform nonparametric estimation. Through a series of numerical simulations under different scenarios, we observed that the proposed approach provides accurate estimates, both in models without jumps and in those involving jump components.

It should be noted that when the jump sizes \(z_i\) are scaled by a multiplicative parameter \(\gamma\), the process defined by \(\gamma z_i\) is indistinguishable from one in which \(\gamma = 1\) and the jump values are replaced by \(z_i' = \gamma z_i\). In other words, the same process can be represented by at least two different parameterizations, making \(\gamma\) non-identifiable unless restrictions are imposed. A common strategy to address this issue is to assume that the jump sizes \(z_i\) are standardized, meaning they have zero mean and unit variance.

Another challenge arises when attempting to estimate a general function \(\gamma(x,z)\) via neural networks. In such cases, the estimation may require evaluating higher-order moments of the jump distribution, which may not exist, depending on the model assumptions (see assumptions \textbf{(A-2)}–\textbf{(A-4)} in \cite{Kumar}).
 
An important direction for future work is the extension of the proposed methodology to multidimensional systems. The approximation scheme developed in this work is, in principle, applicable to SDEs in \(\mathbb{R}^d\), following the multidimensional approximation framework presented in \cite{Kumar}, but its implementation requires careful adaptation of both the numerical scheme and the neural network architecture.

Finally, the development of strategies for the selection of hyper-parameters \((R_1, R_2, R_3, R_4, \text{train}_f)\) remains an open problem. At present, these values are chosen empirically, but automated or adaptive selection methods could significantly enhance both estimation accuracy and training efficiency.

% \begin{acknowledgments}
% % We wish to acknowledge the support of the author community in using
% % REV\TeX{}, offering suggestions and encouragement, testing new versions,
% % \dots.
% \end{acknowledgments}

\section*{Data Availability Statement}

The code and simulated data used for the neural network inference methods presented in this paper are available at: \url{https://github.com/joseramirezgonzalez/NN_levy_jumping}.

\appendix

\section{Characteristic Function}

In this section, we derive analytical expressions for the characteristic function of the random variable \( X_{t+\Delta t} \) conditioned on the sigma-algebra \( \mathcal{F}(X_t) \). These expressions are instrumental in constructing an approximation for the conditional density function \( f_{t,\Delta t|X_t}^{M,h,a} \), introduced in equation~(\ref{fourierden}), via Fourier inversion techniques.
\subsection{Without Additive Jumps}

We begin by recalling a classical result concerning the characteristic function of a general quadratic form in a multivariate Gaussian random vector, which will be central in our derivation.

\begin{lemma}\label{Lem1}
Let \( Z \sim N_k(0, \Sigma) \), and define \( Q := Z'AZ + a'Z + d \), where \( A \in \mathbb{R}^{k \times k} \) is symmetric, \( a \in \mathbb{R}^k \), and \( d \in \mathbb{R} \). Then, the characteristic function of \( Q \) is given by
\begin{equation}
    \phi_Q(u) = |I - 2iuA\Sigma|^{-\frac{1}{2}} e^{iud - \frac{u^2}{2} a' \Sigma^{1/2} \left(I - 2iu \Sigma^{1/2} A \Sigma^{1/2}\right)^{-1} \Sigma^{1/2} a}.
\end{equation}
\end{lemma}

The proof of Lemma~\ref{Lem1} can be found in Theorem 3.2a.3 of \cite{MathaiProvost1992}.

We now apply this result to the setting in which no additive jumps occur, that is, when either \( \lambda = 0 \) or \( \gamma = 0 \). In this case, the dynamics of \( X_{t+\Delta t} \) are governed solely by the drift and diffusion components, and it follows from equation~(\ref{conmil2}) together with Lemma~\ref{Lem1} that
\begin{multline}\label{eq:expectationantes}
E\left( e^{iuX_{t+\Delta t}} \,\big|\, \mathcal{F}(X_t) \right) \\
= e^{iu\left(X_t + \left(f^{\Delta t}(X_t) - \int_{\mathbb{Z}}\gamma(X_t,z)v(dz) - \tfrac{1}{2}g(X_t)g'(X_t)\right)\Delta t \right)} \\
\cdot\ E\left( e^{iu\left(g(X_t)\Delta W_t + \tfrac{1}{2}g(X_t)g'(X_t)(\Delta W_t)^2\right)} \,\Big|\, \mathcal{F}(X_t) \right) \\
= \frac{
e^{iu\left(X_t + \left(f^{\Delta t}(X_t) - \int_{\mathbb{Z}}\gamma(X_t,z)v(dz) - \tfrac{1}{2}g(X_t)g'(X_t)\right)\Delta t \right)
- \frac{\tfrac{1}{2}u^2g(X_t)^2\Delta t}{1 - iug(X_t)g'(X_t)\Delta t}}
}{
\sqrt{1 - iug(X_t)g'(X_t)\Delta t}
}
\end{multline}

\subsection{With Additive Jumping}

We now consider the case where both \( \gamma \neq 0 \) and \( \lambda \neq 0 \). Let us define the sigma-algebras
\begin{multline*}
\mathcal{W}_s^t := \mathcal{F}((W_u)_{s \leq u \leq t}) \quad \text{and} \\
\mathcal{N}_s^t := \mathcal{F}\left((N((s,u], \mathbb{Z}))_{s \leq u \leq t}, (z_i^{(s,t)})_{i=1}^{N((s,t], \mathbb{Z})}\right)
\end{multline*}
By the tower property of conditional expectation, we obtain:\small
\begin{multline}\label{eq:expectation}
E\left(e^{iuX_{t+\Delta t}} \,\big|\, \mathcal{F}(X_t)\right) 
= E\left(E\left(e^{iuX_{t+\Delta t}} \,\big|\, \mathcal{N}_t^{t+\Delta t} \vee \mathcal{F}(X_t)\right) \,\Big|\, \mathcal{F}(X_t) \right) \\
= e^{iu\left(X_t + \left(f^{\Delta t}(X_t)  - \tfrac{1}{2}g(X_t)g'(X_t)\right)\Delta t \right)} \\
\cdot\ E\Bigg( e^{iu\gamma\sum_{i=1}^{N((t,t+\Delta t],\mathbb{Z})}z^{(t,t+\Delta t)}_i} 
\cdot\ E\Bigg( e^{iu\left( g(X_t)\Delta W_t + \tfrac{1}{2} g(X_t)g'(X_t)(\Delta W_t)^2 \right)} \\
\cdot\ e^{iu \sum_{i=1}^{N((t,t+\Delta t],\mathbb{Z})} (g(X_t + \gamma z_i) - g(X_t))(W_{t+\Delta t} - W_{\tau_i})} \,\Big|\, \mathcal{N}_t^{t+\Delta t} \vee \mathcal{F}(X_t) \Bigg) \,\Big|\, \mathcal{F}(X_t) \Bigg).
\end{multline}
\normalsize

We now focus on the inner conditional expectation:
\begin{multline}\label{termino1}
E\Bigg( e^{iu\left( g(X_t)\Delta W_t + \tfrac{1}{2} g(X_t)g'(X_t)(\Delta W_t)^2 \right)} \\
\cdot\ e^{iu \sum_{i=1}^{N((t,t+\Delta t],\mathbb{Z})}(g(X_t + \gamma z_i) - g(X_t))(W_{t+\Delta t} - W_{\tau_i})} \,\Big|\, \mathcal{N}_t^{t+\Delta t} \vee \mathcal{F}(X_t) \Bigg)
\end{multline}

Assume that \( \tau_0 = t \) and suppose \( N((t,t+\Delta t], \mathbb{Z}) = n \) for some \( n \in \mathbb{N} \cup \{0\} \). Then, the jump times \( (\tau_1, \ldots, \tau_n) \) are distributed as the order statistics of \( n \) independent random variables uniformly distributed over \( (t, t+\Delta t) \). We define \( \tau_{n+1} := t+\Delta t \) and set \( z_0 := 0 \). Using this notation, we rewrite the stochastic terms as:
\begin{multline}\label{independent}
g(X_t)\Delta W_t + \tfrac{1}{2} g(X_t)g'(X_t)(\Delta W_t)^2 \\
+ \sum_{i=1}^{n} (g(X_t + \gamma z_i) - g(X_t))(W_{\tau_{n+1}} - W_{\tau_i}) \\
= g(X_t) \sum_{i=1}^{n+1}(W_{\tau_i} - W_{\tau_{i-1}})\\
+ \tfrac{1}{2}g(X_t)g'(X_t)\sum_{i=1}^{n+1}\sum_{j=1}^{n+1}(W_{\tau_i} - W_{\tau_{i-1}})(W_{\tau_j} - W_{\tau_{j-1}}) \\
+ \sum_{i=0}^{n} \sum_{j=i+1}^{n+1} (g(X_t + \gamma z_i) - g(X_t))(W_{\tau_j} - W_{\tau_{j-1}}) \\
= \sum_{i=1}^{n+1}(W_{\tau_i} - W_{\tau_{i-1}}) \left( g(X_t) + \sum_{j=0}^{i-1}(g(X_t + \gamma z_j) - g(X_t)) \right) \\
+ \tfrac{1}{2}g(X_t)g'(X_t)\sum_{i=1}^{n+1}\sum_{j=1}^{n+1}(W_{\tau_i} - W_{\tau_{i-1}})(W_{\tau_j} - W_{\tau_{j-1}}).
\end{multline}
Let \( c_i = g(X_t) + \sum_{j=0}^{i-1} \left(g(X_t + \gamma z^{(t, t + \Delta t)}_j) - g(X_t)\right) \), and let \( c = \frac{1}{2} g(X_t) g'(X_t) \). Using equation~(\ref{independent}) and the independence of Brownian increments, the following expression holds:
\begin{multline}\label{independent23}
Q_{(\tau_1,\dots,\tau_n)}^{(z_1^{(t,t+\Delta t)},\dots,z_n^{(t,t+\Delta t)})} := Z'c\mathbb{J}Z + a'Z + d \\
:= g(X_t)\Delta W_t + \frac{1}{2} g(X_t) g'(X_t) (\Delta W_t)^2 \\
+ \sum_{i=1}^{n}(g(X_t+\gamma z^{(t,t+\Delta t)}_i)-g(X_t))(W_{\tau_{n+1}}-W_{\tau_i}) + \gamma\sum_{i=1}^{n}z^{(t,t+\Delta t)}_i,
\end{multline}
where \( \mathbb{J} \) is the matrix of ones, \( a = (c_1, \dots, c_{n+1}) \), \( d = \gamma \sum_{i=1}^{n} z^{(t,t+\Delta t)}_i \), and \( Z \sim N(0, \Sigma) \) with \( \Sigma_{i,j} = \delta_{i,j}(\tau_i - \tau_{i-1}) \). Using the independence of \( \mathcal{W}_t^{t+\Delta t} \) and \( \mathcal{N}_t^{t+\Delta t} \vee \mathcal{F}(X_t) \), and applying Lemma~\ref{Lem1}, we obtain that the term in equation~(\ref{termino1}) is equal to:
\[
|I - 2iuc \mathbb{J} \Sigma|^{-1/2} \cdot e^{-\frac{u^2}{2} a' \Sigma^{1/2} (I - 2iuc \Sigma^{1/2} \mathbb{J} \Sigma^{1/2})^{-1} \Sigma^{1/2} a}.
\]
Furthermore,
\begin{multline*}
|I - 2iuc \mathbb{J} \Sigma|^{-1/2} = |I - 2iuc \Sigma^{1/2} \mathbb{J} \Sigma^{1/2}|^{-1/2} = |I - 2iuc v v'|^{-1/2}\\ = (1 - 2iuc \sum_{j=1}^{n+1} (\tau_j - \tau_{j-1}))^{-1/2} = \frac{1}{\sqrt{1 - 2iuc \Delta t}},
\end{multline*}
where \( v = (\sqrt{\tau_1 - \tau_0}, \dots, \sqrt{\tau_{n+1} - \tau_n}) \). This follows since the eigenvalues of \( v v' \) are \( \Delta t \) and \( 0 \) (multiplicity \( n \)).

By the Sherman-Morrison formula,
\begin{multline*}
\left(I - 2iuc \Sigma^{1/2} \mathbb{J} \Sigma^{1/2}\right)^{-1} 
= \frac{1}{2iuc} \left( \frac{1}{2iuc} I - v v' \right)^{-1} \\
= I + \frac{2iuc}{1 - 2iuc \Delta t} v v'.
\end{multline*}
Therefore, conditionally on \( \mathcal{N}_t^{t+\Delta t} \vee \mathcal{F}(X_t) \) and given \( N((t,t+\Delta t],\mathbb{Z}) = n \), the characteristic function of $Q_{(\tau_1,\dots,\tau_n)}^{(z_1,\dots,z_n)}$ is given by:
\begin{multline}\label{equationQ*}
\phi_{Q_{(\tau_1,\dots,\tau_n)}^{(z_1,\dots,z_n)}}(u) := \frac{1}{\sqrt{1 - 2iuc \Delta t}} \\
\cdot e^{- \frac{u^2}{2} a' \Sigma^{1/2} \left( I - \frac{4c^2 u^2 \Delta t}{1 + 4c^2 u^2 (\Delta t)^2} v v' + \frac{2icu}{1 + 4c^2 u^2 (\Delta t)^2} v v' \right) \Sigma^{1/2} a + iud}.
\end{multline}
Finally, from equations~(\ref{eq:expectation}) and~(\ref{equationQ*}), it follows that
\begin{multline}\label{laplaceini1}
E\left(  e^{iu X_{t+\Delta t}} \,|\, \mathcal{F}(X_t) \right) = \frac{e^{iu(X_t + (f^{\Delta t}(X_t) - c)\Delta t)}}{\sqrt{1 - 2iuc \Delta t}} \\
\cdot \sum_{n=0}^{\infty} \int_{\mathbb{Z}^n} \int_{t \leq \tau_1 \leq \dots \leq \tau_n \leq t + \Delta t} E\left( \phi_{Q_{(\tau_1,\dots,\tau_n)}^{(z_1,\dots,z_n)}}(u) \,\big|\, \mathcal{F}(X_t) \right) \\
\cdot \frac{n!}{(\Delta t)^n} d\tau_1 \dots d\tau_n \, dF_z(z_1)\dots dF_z(z_n) \, e^{-\lambda \Delta t} \frac{(\lambda \Delta t)^n}{n!},
\end{multline}
where \( F_z \) denotes the distribution of jump sizes.

\section{Density estimation}

In this section, we present numerical studies for the estimation of the approximated density function \( f_{t,\Delta t|X_t}^{M,h,a} \) defined in equation~(\ref{fourierden}). This function is relevant for estimating the drift coefficient \( f \) and the diffusion coefficient \( g \) according to the methodology described in Section~\ref{sec2}, as well as for estimating the parameters \( \gamma \) and \( \lambda \), as detailed in Appendix~C.

When either parameter $\gamma$ or $\lambda$ is equal to zero, the additive jump component in equation~(\ref{equation_red_1}) is not present. In this case, the distribution of $X_{t+\Delta t}$ given the $\sigma$-algebra $\mathcal{F}(X_t)$, according to equation~(\ref{conmil2}), follows the generalized chi-squared distribution. This distribution arises when considering quadratic forms of multivariate normal variables, typically expressed as \( Q = \mathbf{X}^\top A \mathbf{X} + \mathbf{b}^\top \mathbf{X} + c \), where \( \mathbf{X} \sim \mathcal{N}(\boldsymbol{\mu}, \Sigma) \). It generalizes the classical chi-squared distribution by allowing non-zero means, general covariance structures, and additional linear and constant terms. Its cumulative distribution function does not have a closed-form expression. Imhof~\cite{imhof1961computing} proposed a numerical method based on the inversion of the characteristic function. Jensen~\cite{jensen1995saddlepoint} developed saddlepoint approximations specifically for such quadratic forms, with numerical evaluation against Imhof's method. Lindsay et al.~\cite{lindsay2000moment} introduced moment-based mixture approximations and demonstrated their usefulness in applications such as goodness-of-fit and score testing. These methods are implemented in the R package \texttt{CompQuadForm}, which allows for practical computation of distribution functions and $p$-values for generalized chi-squared statistics.

In the general case, for arbitrary values of $\gamma$ and $\lambda$, it can be shown from equation~(\ref{equationQ*}) that the characteristic function of $X_{t+\Delta t}$ given $\mathcal{F}(X_t)$ decays as \( O(1/\sqrt{u}) \) when \( u \to \infty \), and therefore it does not belong to \(  L^2(\mathbb{R}) \). As a result, the existence of a density function cannot be guaranteed (see, e.g.,Lemma 1.1 in ~\cite{fournier2010absolute}).

According to Theorem 6 in~\cite{Yang}, the approximated density function \( f_{t,\Delta t|X_t}^{M,h,a} \)  in equation~(\ref{conmil2}) is defined by (\ref{fourierden}). Theorem~6 in~\cite{Yang} establishes that, under suitable regularity conditions, this approximation converges to the true density—if it exists—as \( Mh \to \infty \) and \( h \to 0 \).

\begin{figure}[ht]
    \centering
    % Primera imagen
    \includegraphics[width=9cm,height=4cm]{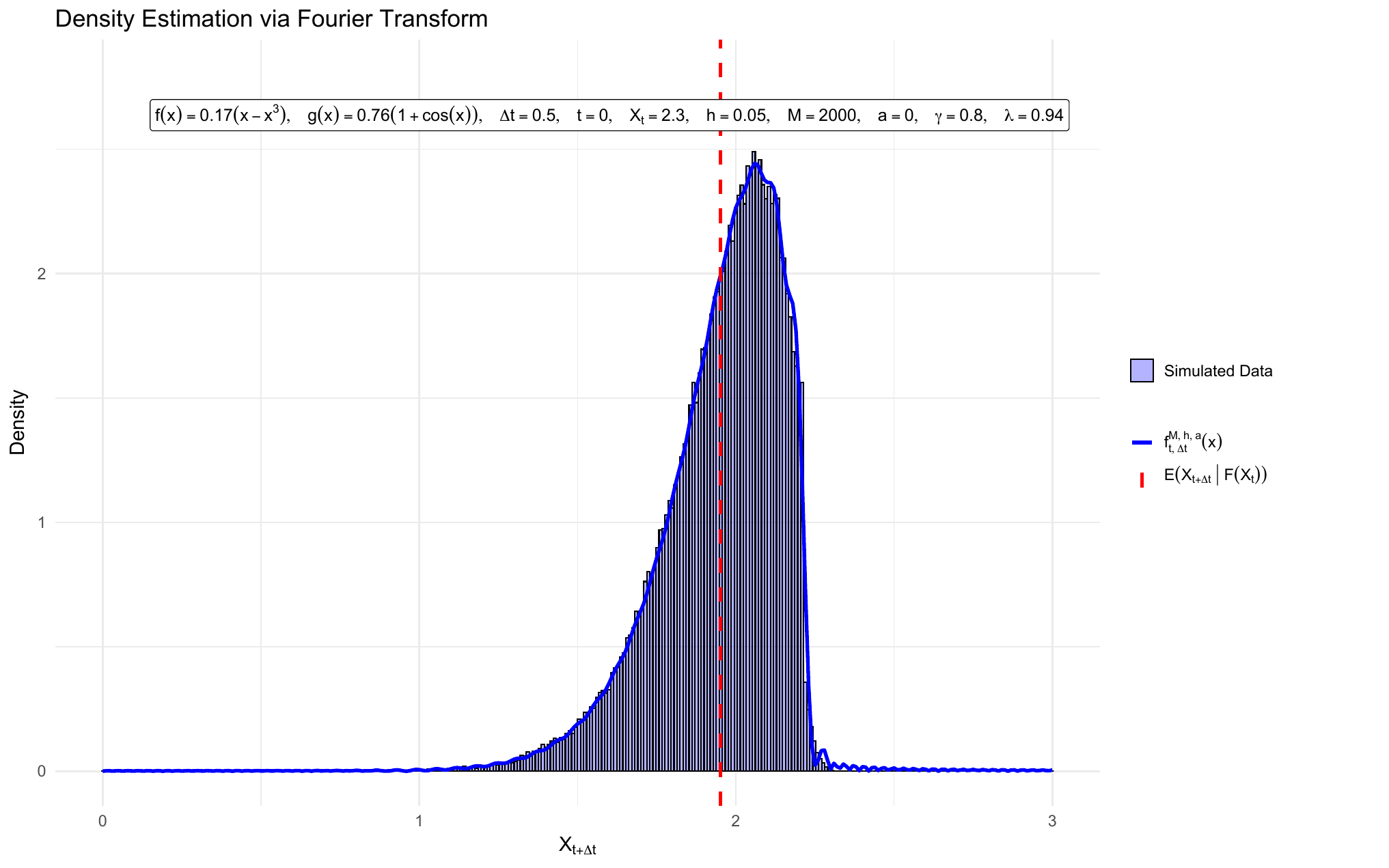}
    \caption{Approximation of the probability density function given by equation~(\ref{fourierden}) using Monte Carlo integration to estimate the integral in equation~(\ref{laplaceini1}). The drift and diffusion coefficients are \(f(x) = 0.17(x - x^3)\) and \(g(x) = 0.76(1 + \cos(x))\), with parameters \(t = 0\), \(\Delta t = 0.5\), \(X_t = 2.3\), \(\gamma = 0.8\), \(\lambda = 0.94\), and jump sizes \(z_i \sim U(-0.1, 0.1)\). The density is approximated using \(M = 2000\), \(h = 0.05\), and \(a = 0\), and is plotted alongside the histogram of 100{,}000 simulated values from equation~(\ref{conmil2}). 
}
    \label{fig:ejemplo1}
\end{figure}

Given equation (\ref{fourierden}), we aim to estimate its probability density function. In Figure \ref{fig:ejemplo1}, we consider the following parameters: $f(x) = 0.17\cdot(x-x^3)$, $g(x) = 0.76\cdot(1+\cos(x))$, $t = 0$, $\Delta t = 0.5$, $X_t =2.3$, $\gamma = 0.8$, $\lambda=0.94$, and we assume that the random variables $z_i$ follow a $U(-0.1,0.1)$ distribution. To estimate the density, we employ the characteristic function . Specifically, we use Monte Carlo integration to approximate the integral given in equation (\ref{laplaceini1}). For this purpose, we generate a sample of 150 simulations of the random variables $N((t, t+\Delta t], Z)$ and $(\tau_1, \ldots, \tau_{N((t, t+\Delta t], Z)})$. 

To estimate the density, we consider equation (\ref{fourierden}) with $M = 2000$, $h = 0.05$, and $a = 0$. Additionally, we perform 100,000 simulations of the random variable in equation (\ref{conmil2}) and plot the approximated density along with the histogram of the simulated data. We observe that the fit between the approximated density and the histogram is satisfactory, particularly near the mode.

\begin{figure}[ht]
    \centering
    \includegraphics[width=9cm,height=4cm]{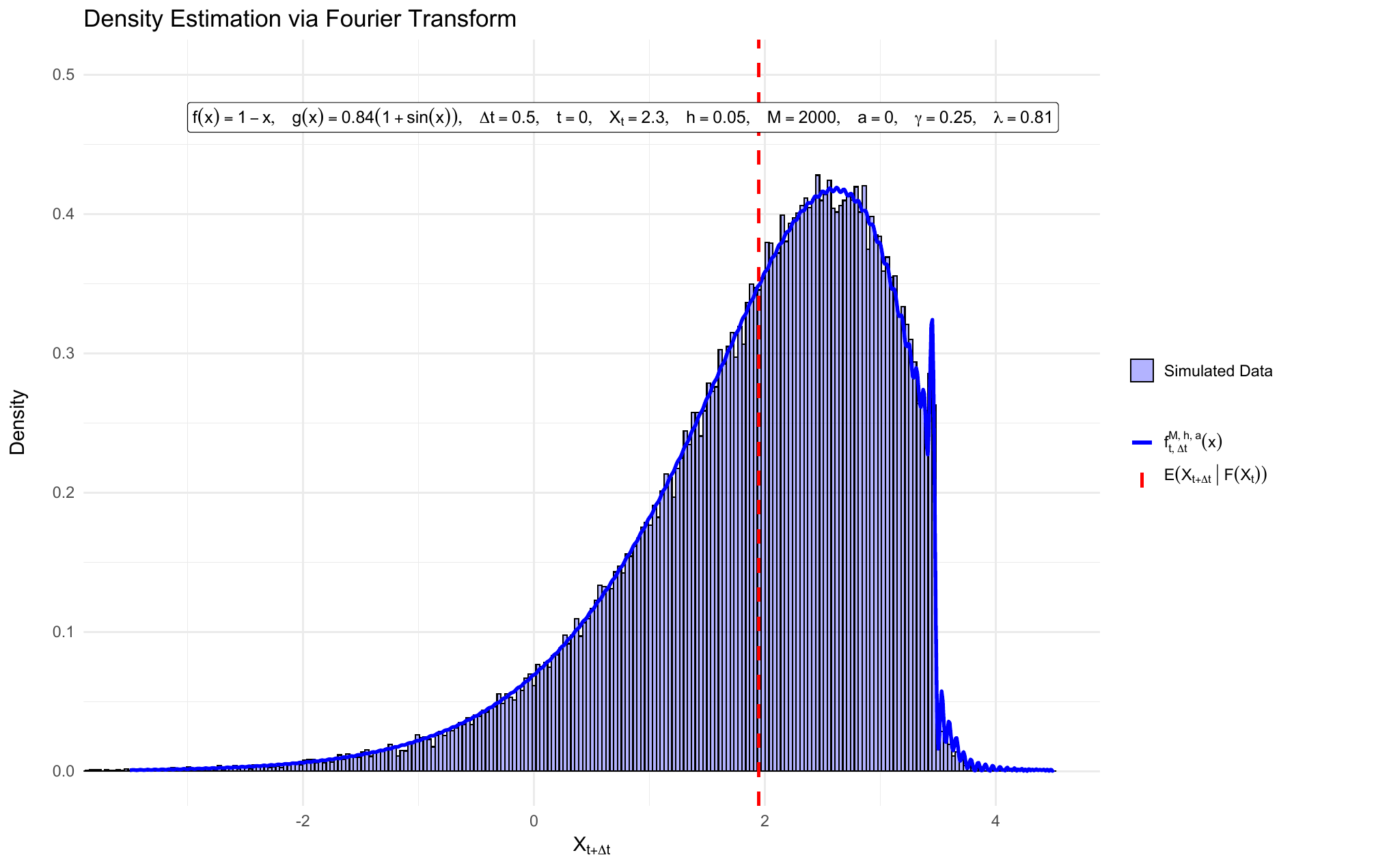}
    \caption{Approximation of the probability density function given by equation~(\ref{fourierden}) using Monte Carlo integration to estimate the integral in equation~(\ref{laplaceini1}). The drift and diffusion coefficients are \(f(x) = 1-x\) and \(g(x) = 0.84(1 + \sin(x))\), with parameters \(t = 0\), \(\Delta t = 0.5\), \(X_t = 2.3\), \(\gamma = 0.25\), \(\lambda = 0.81\), and jump sizes \(z_i \sim N(0, 1)\). The density is approximated using \(M = 2000\), \(h = 0.05\), and \(a = 0\), and is plotted alongside the histogram of 100{,}000 simulated values from equation~(\ref{conmil2}). }
    \label{fig:ejemplo2}
\end{figure}

In Figure \ref{fig:ejemplo2}, we consider the following parameters: $f(x) = 1-x$, $g(x) = 0.84\cdot(1+\sin(x))$, $t = 0$, $\Delta t = 0.5$, $X_t =2.3$, $\gamma = 0.25$, $\lambda=0.81$, and we assume that the random variables $z_i$ follow a $N(0,1)$ distribution. To estimate the density, we employ the characteristic function. Using Monte Carlo integration, we simulate 150 realizations of the random variables $N((t, t+\Delta], Z)$, $(\tau_1, \ldots, \tau_{N((t, t+\Delta], Z)})$ and $(z_1, \ldots, z_{N((t, t+\Delta], Z)})$. For the density estimation, we rely on equation (\ref{fourierden}) with $M = 2000$, $h = 0.05$, and $a = 0$. Again, 100,000 simulations of the random variable from equation (\ref{conmil2}) are performed, and the resulting histogram is plotted alongside the approximated density. The comparison reveals that the estimated density closely aligns with the histogram, indicating a good fit.

\section{Estimation of parameter $\gamma$ and $\lambda$.}\label{parameter_estimation}

Inference for L\'evy processes has been extensively studied using parametric, nonparametric, spectral, and Bayesian methodologies. In the parametric setting, Masuda~\cite{masuda2015parametric} provides a comprehensive overview of estimation techniques for discretely observed L\'evy processes, focusing on quasi-likelihood methods. For nonparametric inference, Comte and Genon-Catalot~\cite{comte2009nonparametric} propose kernel-type estimators for the L\'evy density using high-frequency data under a pure-jump framework. Figueroa-L\'opez and Houdr\'e~\cite{figueroa2006risk} derive non-asymptotic risk bounds for L\'evy density estimators via penalized model selection, while Figueroa-L\'opez~\cite{figueroa2011sieve} constructs sieve-based estimators with provable confidence bands for the L\'evy density. Belomestny~\cite{belomestny2010spectral} introduces a spectral method to estimate the fractional index of a L\'evy process from the empirical characteristic function, and Belomestny and Reiß~\cite{belomestny2015fourier} develop Fourier-based estimators for the L\'evy density using regularized inversion of the characteristic function. In the Bayesian setting, Jasra et al.~\cite{jasra2019bayesian} develop a quasi-likelihood-based MCMC method for stable Lévy-driven SDEs under high-frequency observation. They prove posterior consistency and asymptotic normality via a Bernstein–von Mises theorem and propose a scaling strategy to ensure algorithmic efficiency as data frequency increases. Additionally, Belomestny et al.~\cite{belomestny2019gamma} propose a nonparametric Bayesian method to estimate the Lévy density of infinite activity subordinators using low-frequency data. Their approach combines data augmentation via Gamma process bridges with an infinite-dimensional reversible jump MCMC algorithm. The method achieves posterior consistency and performs well on both simulated and real insurance datasets. These works provide theoretical and computational tools for inference in a broad class of L\'evy-driven models.

Despite the extensive development of estimation techniques for L\'evy processes, we focus on a specific setting where the jump component is modeled as additive symmetric noise driven by a Poisson random measure, according to the approximation in equation~(\ref{conmil2}). In this context, we propose two methods for estimating the parameters $\lambda$ and $\gamma$, tailored to the structure of the approximation.

We estimate the parameters $\lambda$ and $\gamma$, which correspond to the additive symmetric jump noise, as defined in equation~(\ref{conmil2}). Both estimation procedures are based on variants of the Metropolis–Hastings algorithm. The first uses a likelihood ratio, while the second relies on a discriminant function derived from the conditional second moment of $X_{t+\Delta t}$ given $\mathcal{F}(X_t)$, as expressed in equation~(\ref{secondjump}).

Suppose the value of \( X_t \) is known and that we have access to \( N \) independent simulations of the random variable \( X_{t+\Delta t} \), generated according to equation~(\ref{conmil2}). We denote these simulations by \( (x_{t+\Delta t}^{(1)}, \ldots, x_{t+\Delta t}^{(N)}) \). We present both algorithms for parameter estimation:

\subsection{Algorithm 1} To estimate the parameters, we employ the Metropolis–Hastings algorithm using the approximate likelihood function defined in equation~(\ref{fourierden}). Specifically, let
\begin{equation}\label{likelihood_aprox2}
 L_{t,\Delta t}^{M,h,a}(\lambda,\gamma \mid (x_{t+\Delta t}^{(1)},...,x_{t+\Delta t}^{(N)}),f,g) := \prod_{i=1}^{N} f_{t,\Delta t|X_t}^{M,h,a}(x_{t+\Delta t}^{(i)}),
\end{equation}
and define the likelihood ratio as
\begin{equation}
R((\lambda_1,\gamma_1)\mid(\lambda_0,\gamma_0)) := \frac{L_{t,\Delta t}^{M,h,a}(\lambda_1,\gamma_1 \mid (x_{t+\Delta t}^{(1)}, \ldots, x_{t+\Delta t}^{(N)}),f,g)}{L_{t,\Delta t}^{M,h,a}(\lambda_0,\gamma_0 \mid (x_{t+\Delta t}^{(1)}, \ldots, x_{t+\Delta t}^{(N)}),f,g)}.
\end{equation}

The proposal distribution is given by
\begin{equation}
q((\lambda_1,\gamma_1)\mid(\lambda_0,\gamma_0)) = f_{\lambda_0,\sigma_1^2}(\lambda_1) f_{\gamma_0,\sigma_2^2}(\gamma_1),
\end{equation}
where, for \( a, b > 0 \), the function \( f_{a,b}(x) \) denotes the density of a log-normal distribution with parameters \( (\log(a) - \tfrac{b^2}{2}, b^2) \). We present below the algorithm used to simulate a random sample from the posterior distribution of \( (\lambda, \gamma) \) using Metropolis–Hastings.

\refstepcounter{myalgorithm}
\phantomsection
\label{algo1t}
\vspace{0.2em}
\noindent
\rule{\linewidth}{0.4pt}
\vspace{-1.1em}

\begin{center}
\textbf{Algorithm~\themyalgorithm:} Metropolis–Hastings with Approximated Likelihood Ratio.
\end{center}

\vspace{-1.1em}
\rule{\linewidth}{0.4pt}
\vspace{-0.7em}

\noindent
\textbf{Input:} approximate likelihood ratio \( R((\lambda_1,\gamma_1)\mid(\lambda_0,\gamma_0)) \) \\
\textbf{Input:} proposal distribution \( q((\lambda_1,\gamma_1)\mid(\lambda_0,\gamma_0)) \) \\
Initialize \( (\lambda,\gamma) \gets (\lambda_0,\gamma_0) \)

\vspace{-0.3em}
\noindent
\textbf{For} \( i = 1 \) to \( m \) \textbf{do:}

\vspace{-0.5em}
\begin{flushleft}
\hspace{1em}Sample \( (\lambda_1,\gamma_1) \sim q((\lambda_1,\gamma_1)\mid(\lambda_0,\gamma_0)) \)\\[0.3em]
\hspace{1em}Compute:
\[
P = \min\left\{1, R((\lambda_1,\gamma_1)\mid(\lambda_0,\gamma_0)) \cdot \frac{q((\lambda_0,\gamma_0)\mid(\lambda_1,\gamma_1))}{q((\lambda_1,\gamma_1)\mid(\lambda_0,\gamma_0))} \right\}
\]
\vspace{-1.5em}
\hspace{1em}Set:
\[
(\lambda^{(i)},\gamma^{(i)}) = 
\begin{cases}
(\lambda_1,\gamma_1), & \text{with probability } P \\
(\lambda_0,\gamma_0), & \text{with probability } 1 - P
\end{cases}
\]
\vspace{-1.2em}
\hspace{1em}Update \( (\lambda_0,\gamma_0) \gets (\lambda^{(i)},\gamma^{(i)}) \)
\end{flushleft}

\vspace{0.4em}
\noindent
\textbf{End for}

\vspace{0.2em}
\noindent
\textbf{Output:} \( \{(\lambda^{(1)},\gamma^{(1)}),\dots,(\lambda^{(m)},\gamma^{(m)})\} \)

\vspace{0.2em}
\noindent
\rule{\linewidth}{0.4pt}

\subsection{Algorithm 2} 
We now present a Metropolis–Hastings simulation algorithm for the parameters \( (\lambda, \gamma) \), based on the conditional second moment of \( X_{t+\Delta t} \) given \( \mathcal{F}(X_t) \). We define \( h^{N}(\lambda,\gamma) := h(\lambda,\gamma \mid (x_{t+\Delta t}^{(1)}, \ldots, x_{t+\Delta t}^{(N)})) \), with
\small
\begin{multline}\label{h_MH}
h(\lambda,\gamma \mid (x_{t+\Delta t}^{(1)}, \ldots, x_{t+\Delta t}^{(N)})) := \frac{1}{N} \sum_{i=1}^{N} \Bigg( \Bigg( x^{(i)}_{t+\Delta t} - X_t - \frac{f(X_t)}{1 + \Delta t f^2(X_t)} \Delta t \Bigg)^2 \\
- \mathbb{E} \left[ \left( X_{t+\Delta t} - \mathbb{E}(X_{t+\Delta t} \mid \mathcal{F}(X_t)) \right)^2 \,\Big|\, \mathcal{F}(X_t) \right] \Bigg)^2
\end{multline}
\normalsize
where the second expectation is evaluated under the true values of \( \lambda \) and \( \gamma \). By equation~(\ref{secondjump}) and the law of large numbers, we obtain \( h^N(\lambda, \gamma) \xrightarrow[N \to \infty]{} 0 \). Therefore, a point estimator can be defined as
\[
(\lambda_{\text{max}}, \gamma_{\text{max}}) = \arg\min_{(\lambda,\gamma)} h^N(\lambda,\gamma).
\]

In addition, for large \( N \), if \( h^N(\lambda_1,\gamma_1) \leq h^N(\lambda_0,\gamma_0) \), we interpret \( (\lambda_1,\gamma_1) \) as more probable than \( (\lambda_0,\gamma_0) \) given the sample. Based on this idea, and following Algorithm 3 of \cite{neklyudov2019implicit}, we propose a Metropolis–Hastings-type algorithm to simulate probable values of \( (\lambda,\gamma) \). We consider a learned discriminator defined as
\begin{equation}
d_{\theta}((\lambda,\gamma)) := \exp\left(\theta \exp(h^N(\lambda,\gamma))\right), \quad \theta > 0.
\end{equation}

We present below the algorithm used to simulate a random sample from the posterior distribution of \( (\lambda, \gamma) \):

\refstepcounter{myalgorithm}
\phantomsection
\label{algo2t}
\vspace{0.2em}
\noindent
\rule{\linewidth}{0.4pt}
\vspace{-1.1em}

\begin{center}
\textbf{Algorithm~\themyalgorithm:} Metropolis–Hastings with Discriminator Based on Conditional Second Moment
\end{center}

\vspace{-1.1em}
\rule{\linewidth}{0.4pt}
\vspace{-0.7em}

\noindent
\textbf{Input:} learned discriminator \( d_\theta((\lambda,\gamma)) \) \\
\textbf{Input:} proposal distribution \( q((\lambda_1,\gamma_1)\mid(\lambda_0,\gamma_0)) \) \\
Initialize \( (\lambda,\gamma) \gets (\lambda_0,\gamma_0) \)

\vspace{-0.3em}
\noindent
\textbf{For} \( i = 1 \) to \( m \) \textbf{do:}

\vspace{-0.5em}
\begin{flushleft}
\hspace{1em}Sample proposal \( (\lambda_1,\gamma_1) \sim q((\lambda_1,\gamma_1)\mid(\lambda_0,\gamma_0)) \)\\[0.3em]
\hspace{1em}Compute:
\[
P = \min\left\{1, \frac{d_\theta((\lambda_1,\gamma_1))}{d_\theta((\lambda_0,\gamma_0))} \right\}
\]
\vspace{-1.5em}
\hspace{1em}Set:
\[
(\lambda^{(i)},\gamma^{(i)}) = 
\begin{cases}
(\lambda_1,\gamma_1), & \text{with probability } P \\
(\lambda_0,\gamma_0), & \text{with probability } 1 - P
\end{cases}
\]
\vspace{-1.2em}
\hspace{1em}Update \( (\lambda_0,\gamma_0) \gets (\lambda^{(i)},\gamma^{(i)}) \)
\end{flushleft}

\vspace{0.4em}
\noindent
\textbf{End for}

\vspace{0.2em}
\noindent
\textbf{Output:} \( \{(\lambda^{(1)},\gamma^{(1)}),\dots,(\lambda^{(m)},\gamma^{(m)})\} \)

\vspace{0.2em}
\noindent
\rule{\linewidth}{0.4pt}

Note that if \( h^N(\lambda_1,\gamma_1) \leq h^N(\lambda_0,\gamma_0) \), then \( \frac{d_\theta((\lambda_1,\gamma_1))}{d_\theta((\lambda_0,\gamma_0))} > 1 \), and consequently \( P = 1 \), implying that the proposed sample is accepted with probability one. The use of a double exponential function in \( d_\theta \) acts as a penalty mechanism for large variations in the value of \( h^N(\lambda,\gamma) \), thereby promoting smoother transitions between successive accepted values.

\subsection{Numerical studies} 

\begin{figure}[ht]
    \centering
    % Subfigura A: histograma
        \includegraphics[width=9cm,height=4cm]{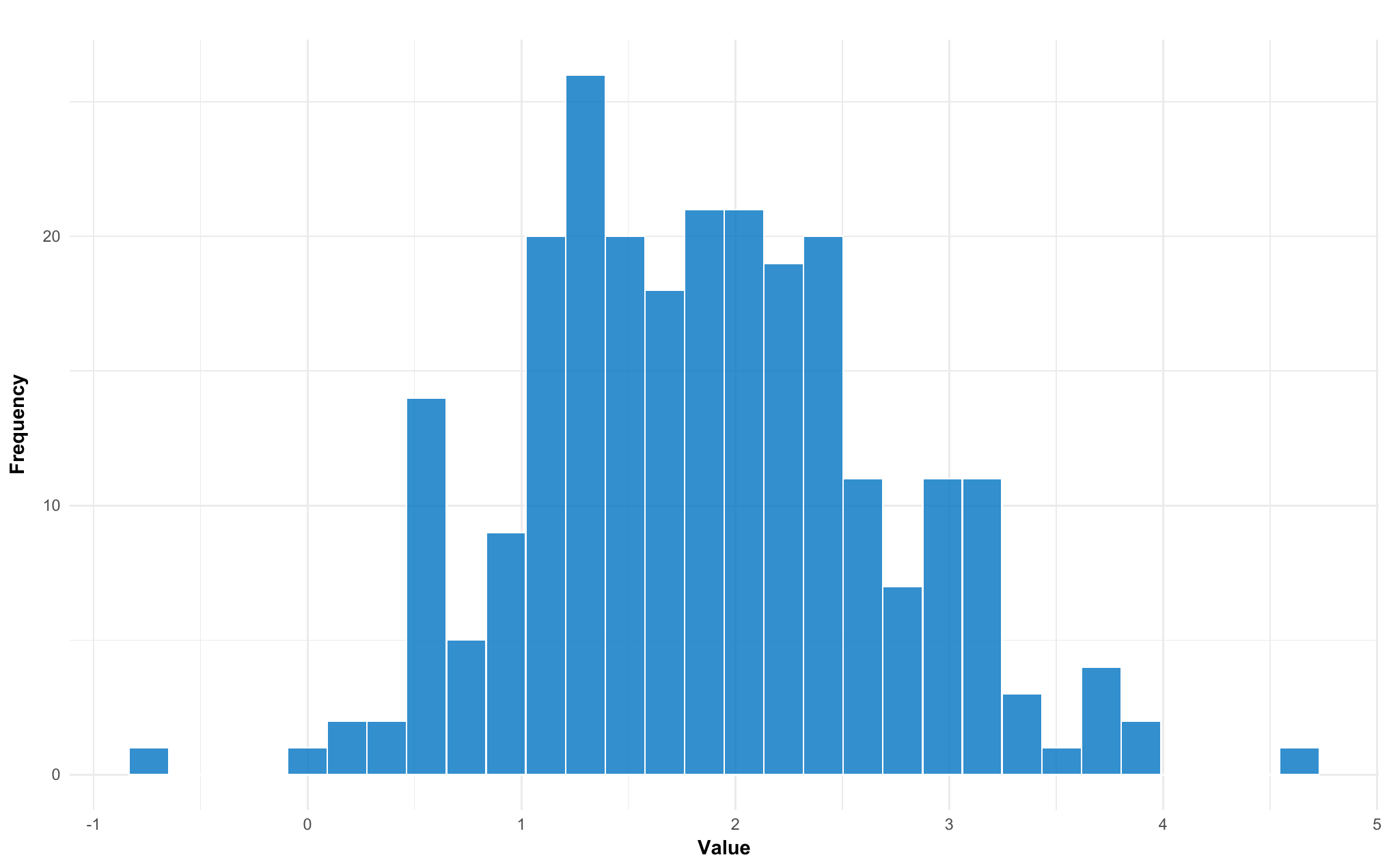}
       \caption{ Histogram of the simulated values of \(X_{t+\Delta t}\) generated from the model defined in equation~(\ref{conmil2}) with \(f(x) = \sin(x)\), \(g(x) = 0.35x + 0.2\), \(\lambda = 1.7\), and \(\gamma = 2.4\). The sample size is \(N = 250\). }
    \label{fig:data_sim_test}
\end{figure}

To test the performance of the proposed algorithms, we generated synthetic data from the model defined in equation~(\ref{conmil2}). We fixed the initial condition \( X_t = 1.5 \), with \( \Delta t = 0.5 \), and \( t = 0 \). The drift and diffusion functions used in the simulation were:
\begin{equation*}
f(x) = \sin(x), \quad g(x) = 0.35x + 0.2,
\end{equation*}
and the jump magnitudes were drawn from a uniform distribution \( \text{Unif}(-0.5, 0.5) \). The parameters are given by \( \lambda = 1.7 \) and \( \gamma = 2.4 \). We generated a sample of size \( N = 250 \) of the variable \( X_{t+\Delta t} \), conditional on the known value of \( X_t \). In Figure~\ref{fig:data_sim_test}, we present a histogram of the simulated sample \( (x_{t+\Delta t}^{(1)}, \ldots, x_{t+\Delta t}^{(N)}) \).

\begin{figure}[ht]
\centering
\includegraphics[width=9cm,height=5cm]{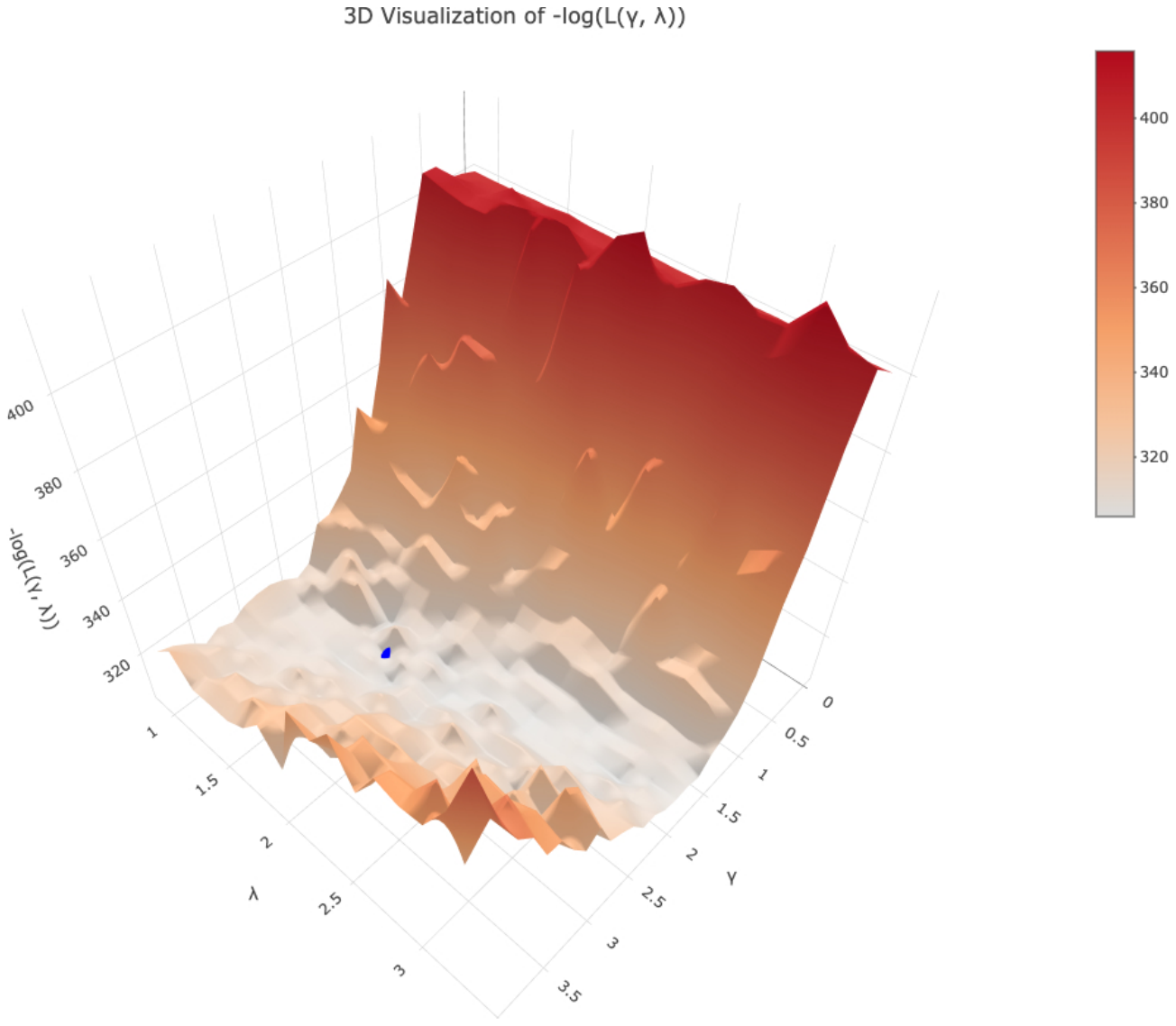}
\par\smallskip
\textit{Panel A: $-\log(L_{t,\Delta t}^{M,h,a}(\lambda,\gamma \mid (x_{t+\Delta t}^{(1)}, \ldots, x_{t+\Delta t}^{(N)}),f,g))$.}
\par\medskip
\includegraphics[width=9cm,height=5.5cm]{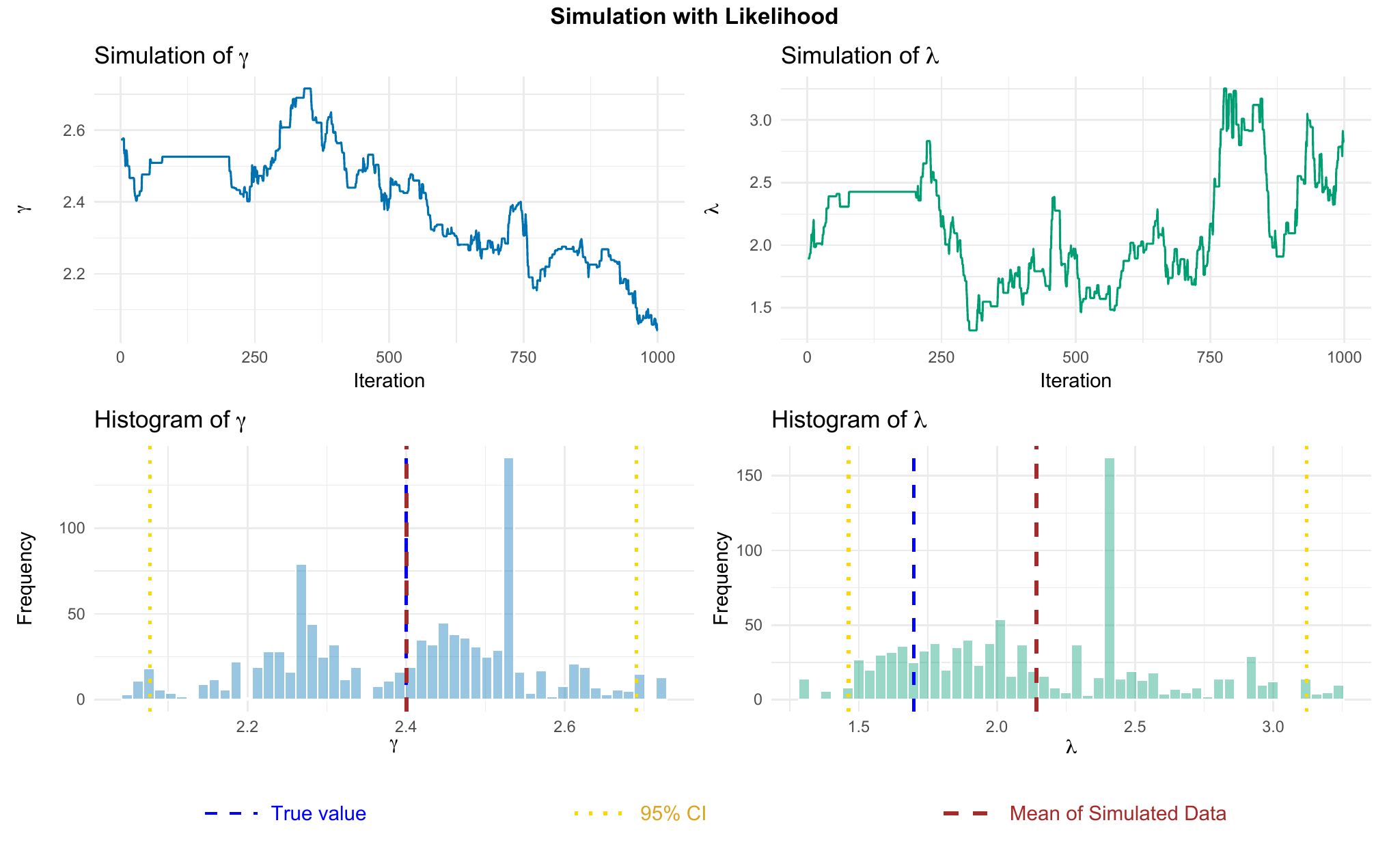}
\par\smallskip
\textit{Panel B: Metropolis-Hastings simulation using Algorithm~\ref{algo1t}.}
\caption{\label{fig:algo1}
Estimation of the parameters \(\lambda\) and \(\gamma\) using Algorithm~\ref{algo1t}. Panel A shows the negative log-likelihood function \(-\log(L_{t,\Delta t}^{M,h,a}(\lambda,\gamma \mid (x_{t+\Delta t}^{(1)}, \ldots, x_{t+\Delta t}^{(N)}),f,g))\), with \(M = 200\), \(h = 0.05\), and \(a = 0\). Panel B displays the histogram of posterior samples obtained from 1000 iterations of the Metropolis–Hastings algorithm with \(\sigma_1 = 0.05\) and \(\sigma_2 = 0.01\).}
\end{figure}

To estimate the parameters \( \lambda \) and \( \gamma \) using Algorithm 1, we employed the approximate likelihood function defined in equation~(\ref{fourierden}) with parameters \( M = 200 \), \( h = 0.05 \), and \( a = 0 \). The characteristic function approximation required for the likelihood computation was evaluated using \( n = 200 \) Monte Carlo samples. The Metropolis–Hastings algorithm was run for \( m = 1000 \) iterations with $\sigma_1=0.05,\phantom{ }\sigma_2=0.01$ in the proposal distribution.  In Figure~\ref{fig:algo1} (A), we present the plot of $-\log(L_{t,\Delta t}^{M,h,a}(\lambda,\gamma \mid (x_{t+\Delta t}^{(1)}, \ldots, x_{t+\Delta t}^{(N)}),f,g))$, and in Figure~\ref{fig:algo1} (B), the histogram of the resulting simulated data is shown using Algorithm~\ref{algo1t}.

The posterior means of the estimated parameters based on the sample are:
\begin{equation*}
\widehat{\lambda}_{\text{mean}} = 2.142847, \quad \widehat{\gamma}_{\text{mean}} = 2.400442,
\end{equation*}
with 95\% confidence intervals (CIs) given by:
\begin{equation*}
\widehat{\lambda} \in [1.463014, 3.120582], \quad \widehat{\gamma} \in [2.076857, 2.690452].
\end{equation*}
The true parameter values lie within the corresponding CIs.

\begin{figure}[ht]
\centering
\includegraphics[width=9cm,height=5cm]{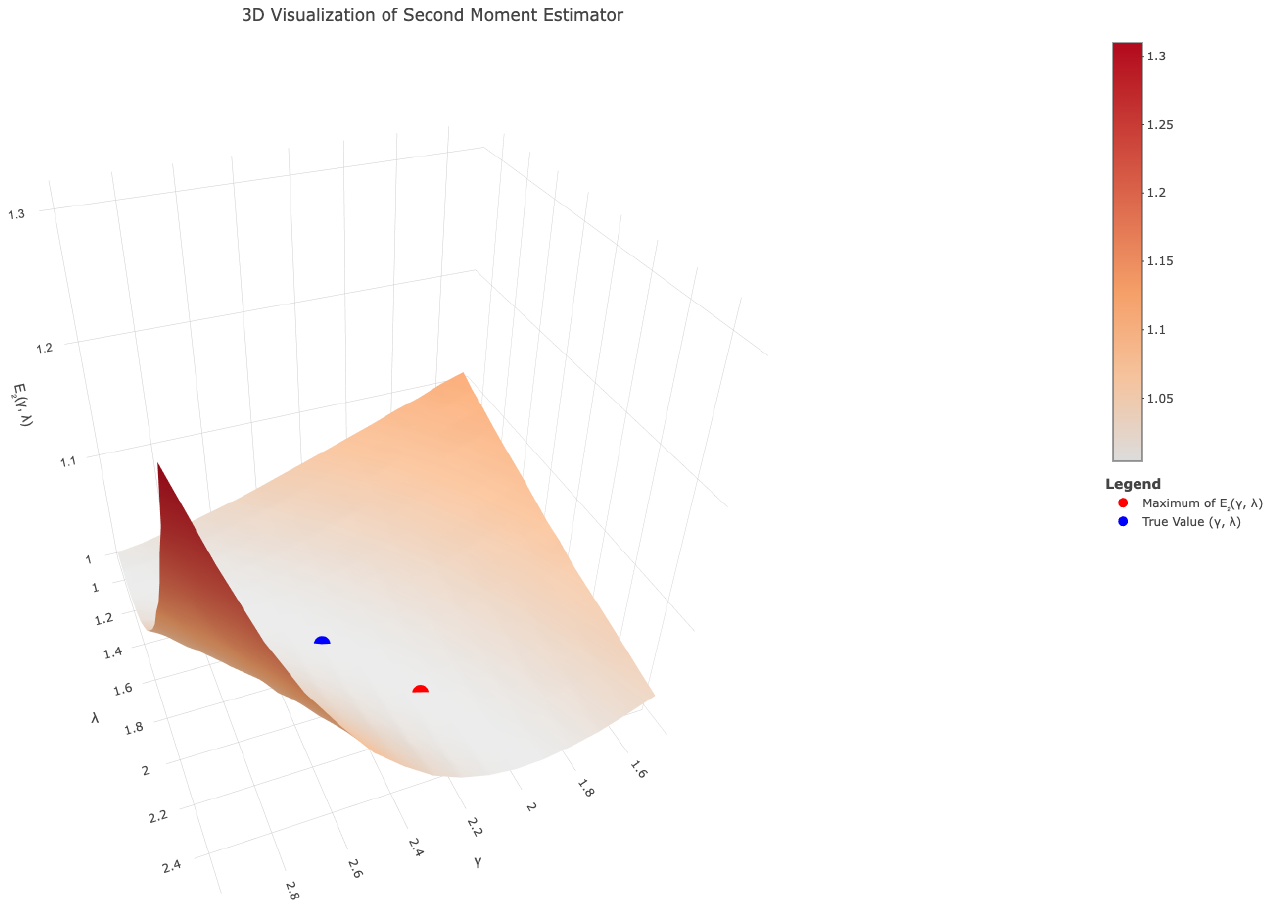}
\par\smallskip
\textit{Panel A: $h^N(\lambda,\gamma)$.}
\par\medskip
\includegraphics[width=9cm,height=5.5cm]{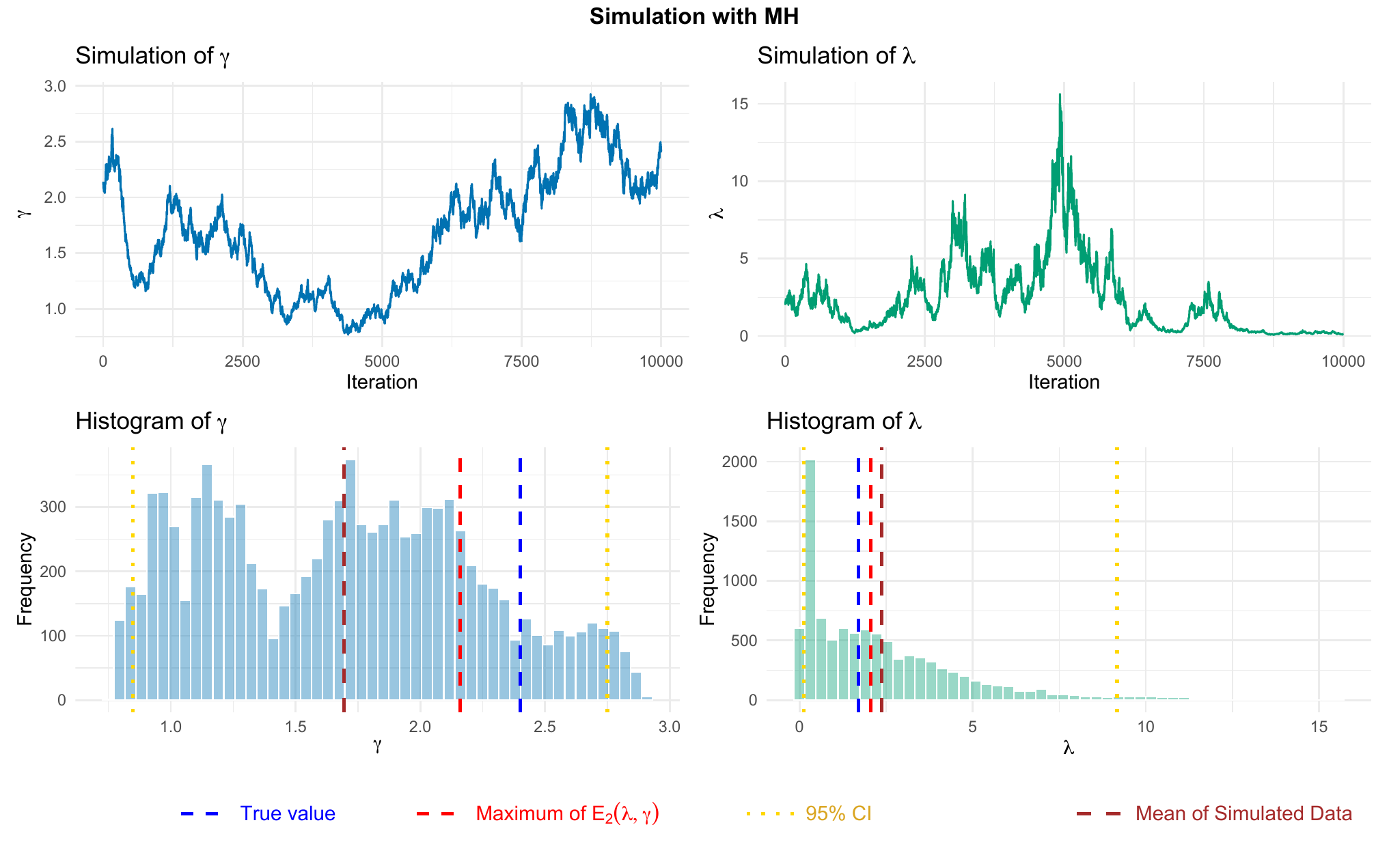}
\par\smallskip
\textit{Panel B: Metropolis-Hastings simulation using Algorithm~\ref{algo2t}.}
\caption{\label{fig:algo2}
Estimation of the parameters \(\lambda\) and \(\gamma\) using Algorithm~\ref{algo2t}. Panel A shows the surface of the objective function \(h^N(\lambda,\gamma)\), minimized using the Nelder–Mead method (red point). Panel B displays the histogram of posterior samples obtained from 10,000 iterations of the Metropolis–Hastings algorithm with \(\sigma_1 = 0.05\) and \(\sigma_2 = 0.01\).}
\end{figure}

We now apply Algorithm 2, which requires an estimate of the second moment defined by equations~(\ref{secondjump}) and~(\ref{h_MH}). This estimate is obtained via simulations and the law of large numbers, using \( n = 4000 \) samples. For the discriminator function \( d_\theta \), we set \( \theta = 5 \). The number of Metropolis–Hastings iterations was fixed at \( m = 10{,}000 \), with proposal distribution parameters \( \sigma_1 = 0.05 \) and \( \sigma_2 = 0.01 \). In Figure~\ref{fig:algo2} (A), we present the plot of $h^{N}(\lambda,\gamma)$. Using the \texttt{optim} function with the "Nelder-Mead" method in R, we minimized the function $h^{N}(\lambda,\gamma)$. The resulting estimate is given by
\[
(2.0656, 2.1592) = \arg\min_{(\lambda,\gamma)} h^N(\lambda,\gamma).
\]
 
In Figure~\ref{fig:algo2} (B), the histogram of the resulting simulated data is shown using Algorithm~\ref{algo2t}. The posterior means of the estimated parameters based on the sample are:
\begin{equation*}
\widehat{\lambda}_{\text{mean}} = 2.380231, \quad \widehat{\gamma}_{\text{mean}} = 1.694045,
\end{equation*}
with corresponding 95\% CIs:
\begin{equation*}
\widehat{\lambda} \in [0.1208828,\ 9.1716519], \quad \widehat{\gamma} \in [0.8473878,\ 2.7503508].
\end{equation*}
The true parameter values used for the data generation, \( \lambda = 1.7 \) and \( \gamma = 2.4 \), lie within the estimated CIs, indicating the consistency of the proposed estimation procedure under this simulation setting.

In both algorithms, the true parameter values lie within a $95\%$ CIs based on the simulated sample. However, the CIs produced by Algorithm~\ref{algo1t} are approximately 5.46 times narrower for $\lambda$ and 3.1 times narrower for $\gamma$ compared to those produced by Algorithm~\ref{algo2t}. Nonetheless, the computational cost of Algorithm~\ref{algo2t} is lower than that of Algorithm~\ref{algo1t}, as the latter requires computing the characteristic function  of $X_{t+\Delta t}$ given $\mathcal{F}(X_t)$ over a grid of points of the form $mh + ia$, with $m = -M, -(M-1), \ldots, M-1, M$. At each grid point, the characteristic function  is estimated via Monte Carlo simulation using either equation~(\ref{laplaceini1}), where each integral evaluation has a computational cost of order $N(\Delta t,\lambda)^3$, with $N(\Delta t,\lambda) \sim \text{Pois}(\lambda \Delta t)$. Subsequently, the approximation in equation~(\ref{likelihood_aprox2}) is applied. In contrast, Algorithm~\ref{algo2t} only requires approximating the expected value in equation~(\ref{secondjump}) using the law of large numbers, which has computational order $N(\Delta t,\lambda)^2$.

Algorithm~\ref{algo1t} achieves significantly tighter CIs, which reflects a higher estimation precision, but this comes at the expense of a substantially greater computational burden. In contrast, Algorithm~\ref{algo2t} is less precise but provides a more computationally efficient alternative, while still capturing the true parameters within the estimated intervals.

\small

\normalsize

\bibliography{aipsamp}

\end{document}